\documentclass{article}

\usepackage{arxiv}
\usepackage[utf8]{inputenc} % allow utf-8 input
\usepackage[T1]{fontenc}    % use 8-bit T1 fonts
\usepackage{hyperref}       % hyperlinks
\usepackage{url}            % simple URL typesetting
\usepackage{booktabs}       % professional-quality tables
\usepackage{amsfonts}       % blackboard math symbols
\usepackage{nicefrac}       % compact symbols for 1/2, etc.
\usepackage{microtype}      % microtypography
\usepackage{lipsum}		% Can be removed after putting your text content
\usepackage{graphicx}
\usepackage{natbib}
\usepackage{doi}
\usepackage{multirow}
\usepackage{authblk}
\usepackage[symbol]{footmisc}
\usepackage{lineno}
\usepackage{tabularx}
\usepackage{xcolor}
\usepackage{marvosym}
\usepackage{ifsym}

\setcitestyle{numbers,square}

\title{Learning a Multi-task Transformer via \\ Unified and Customized Instruction Tuning for \\ Chest Radiograph Interpretation }

\author[$\ast$,\Letter,1,2]{Lijian~Xu}
\author[$\ast$,3]{Ziyu Ni} 
\author[3]{Xinglong~Liu}
\author[\Letter,2]{Xiaosong~Wang}
\author[1,4]{Hongsheng~Li} 
\author[\Letter,2]{Shaoting~Zhang}

\affil[1]{Centre for Perceptual and Interactive Intelligence, the Chinese University of Hong Kong, Hong Kong}
\affil[2]{Shanghai Artificial Intelligence Laboratory, Shanghai}
\affil[3]{Sensetime Research, Shanghai}
\affil[4]{Department of Electronic Engineering, the Chinese University of Hong Kong, Hong Kong}
\affil[$\ast$]{Equal contributions}
\date{}

\renewcommand{\headeright}{Learning a Multi-task Transformer for Chest Radiograph Interpretation}
 \renewcommand{\shorttitle}{}
\begin{document}
\maketitle

\begin{abstract}
The emergence of multi-modal deep learning models has made significant impacts on clinical applications in the last decade. However, the majority of models are limited to single-tasking, without considering disease diagnosis is indeed a multi-task procedure. Here, we demonstrate a unified transformer model specifically designed for multi-modal clinical tasks by incorporating customized instruction tuning. 
% towards a  model for chest radiography.
We first compose a multi-task training dataset comprising 13.4 million instruction and ground-truth pairs (with approximately one million radiographs) for the customized tuning, involving both image- and pixel-level tasks. 
Thus, we can unify the various vision-intensive tasks in a single training framework with homogeneous model inputs and outputs to increase clinical interpretability in one reading. Finally, we demonstrate the overall superior performance of our model compared to prior arts on various chest X-ray benchmarks across multi-tasks in both direct inference and finetuning settings.
Three radiologists further evaluate the generated reports against the recorded ones, which also exhibit the enhanced explainability of our multi-task model. 

%TC:endignore
\end{abstract}
% keywords can be removed

\section{Introduction}

Chest radiography (CXR) is a non-invasive and relatively low-cost diagnostic radiology examination for screening and diagnosis of various thoracic diseases affecting the lung and heart \cite{STEVENS20211006}. However, the interpretation of CXR is greatly challenged by its low sensitivity of subtle abnormalities, overlapping structures, and limited soft tissue details, and therefore, depends heavily on the capability and experience of radiologists \cite{CALLI2021102125}. On the other hand, the growing demand for CXR examination has brought a burden on medical professionals, which also limits the clinical application of CXR, especially in community clinics or primary hospitals. In this context, automated diagnosis by AI could potentially contribute to reducing the workload of radiologists.

Large Language Models ~\cite{dosovitskiy2021image,liu2021swin} 
% have revolutionized natural language processing, and self-attention-based architectures, in particular Transformers~\cite{vaswani2017attention}, have become the model of choice in language and visual tasks~\cite{dosovitskiy2021image,liu2021swin}. By leveraging extensive textual data, these models 
have revolutionized natural language processing and developed the capability to generate responses that closely resemble those from humans. They excel at a wide range of tasks, including language translation, question answering, and text generation \cite{devlin2019bert,touvron2023llama,scao2022bloom,chowdhery2022palm,zhang2022opt,du2022glam,zeng2022glm}. 
Models like ChatGPT (OpenAI) \cite{brown2020language,ouyang2022training} and Med-PaLm (Google) \cite{singhal2022large,singhal2023towards} have also demonstrated powerful reasoning capabilities of language models in complex scenarios like medical diagnosis to assist professionals in delivering care.
% Language models with prompt learning~\cite{brown2020language,ouyang2022training,chen2021evaluating} prove powerful zero-/few-shot learners and provide a new paradigm for human-computer interaction.
% Building upon this, multimodal large models utilize pre-trained image encoder and text encoder then align visual-language features with simple linear layers\cite{radford2021learning,alayrac2022flamingo,li2023blip2,chen2022pix2seq,wang2022ofa,wang2022image,chen2023pali,li2022grounded}. The stronger generalizable language features are leveraged to guide the extraction of visual features. 
% These models demonstrate impressive joint understanding capabilities of language and images, allowing users to give instructions in natural language to perform specific tasks. 
% Multimodal large models can recognize objects in images and accomplish tasks such as Visual Question Answering (VQA) and Image Captioning. 
Nonetheless, such tasks are limited to a more general medical scope and largely rely on the visual features on the image level, without touching the pixel-level vision tasks, e.g., disease localization and segmentation. Moreover, to further improve the model's ability on downstream tasks, supervised instruction tuning with specific downstream task-oriented data is often required on language only \cite{wei2022finetuned,chung2022scaling} and vision-language tasks \cite{dai2023instructblip,zhu2023minigpt4,liu2023visual}, individually.

\textcolor{black}{On the other hand}, the development of the multi-modal model in the medical field has lagged. Most models are designed primarily for "pure" language tasks \cite{rasmy2021med,singhal2022large}. 
Several generalist models for the biomedical field have been recently proposed and achieved progress in the VQA task \cite{moor2023foundation,xu2023elixr,zhang2023biomedgpt,chen2023medblip,tu2023towards,wu2023generalist,Hu_2023,moor2023medflamingo,li2023llavamed}. On the other hand, the present multi-modality models are not well-suited for traditional image processing tasks like detection and segmentation. Existing methods face discrepancies in input, output, and training processes between visual tasks and language tasks, which hinders efficient collaboration. Furthermore, relying solely on textual outputs restricts the answer capacity and interpretability to some extent. For instance, in computer-aided diagnosis 
 using medical images, while the model can identify the disease type and provide treatment recommendations, it is unable to pinpoint the exact location and region of the pathologies, limiting its clinical usefulness as a reference for explainable diagnosis prediction.

\begin{figure}[tbh!]
    \centering
    %\fbox{\rule[-.5cm]{4cm}{4cm} \rule[-.5cm]{4cm}{0cm}}
    \includegraphics[width=1\linewidth]{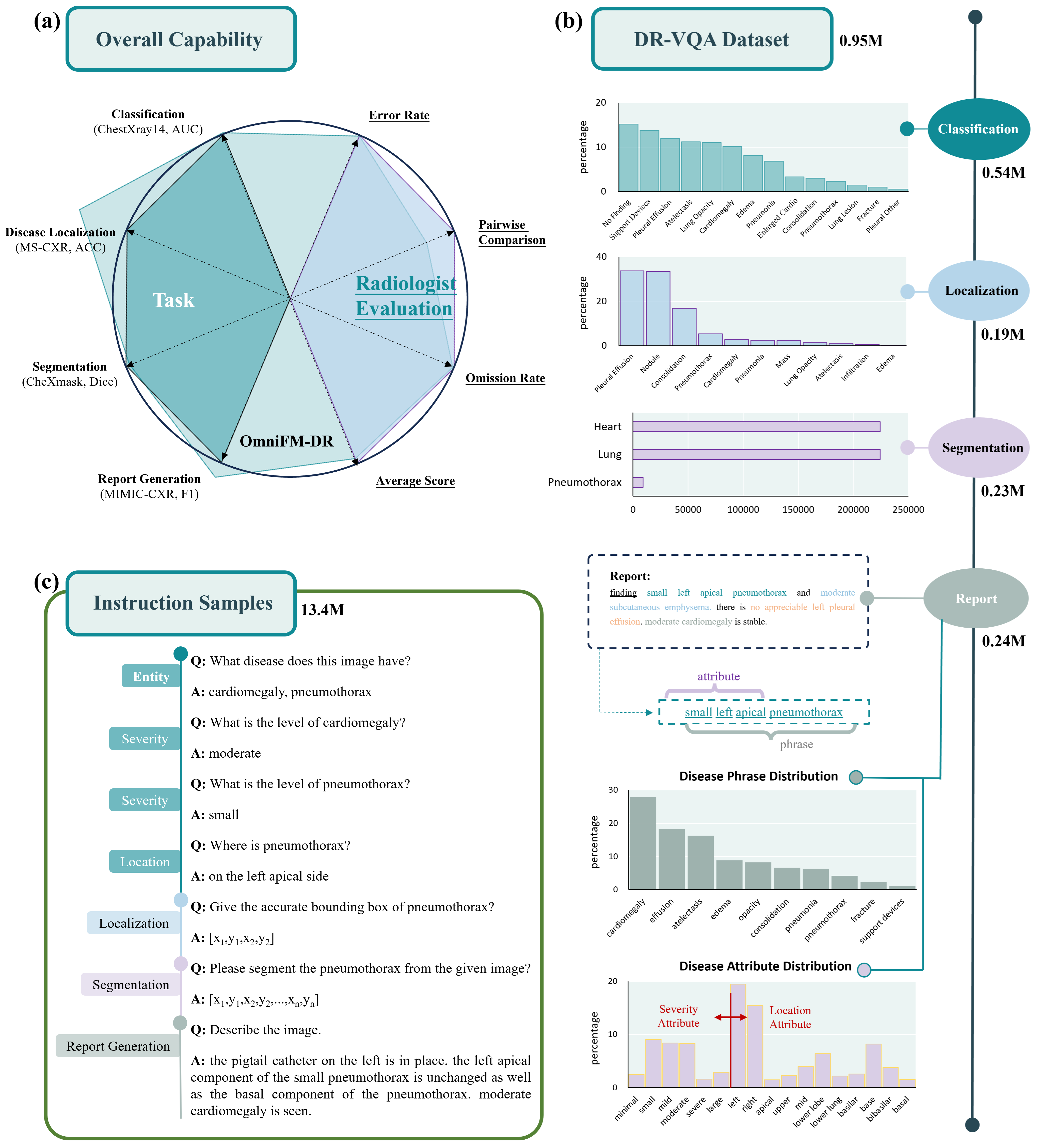}
    \caption{(a) Overview performance of the proposed method \textbf{OmniFM-DR} on multiple datasets across multi-tasks (i.e., disease classification, localization, segmentation, and report generation). \textcolor{black}{OmniFM-DR's metrics exceed the circle indicates better performance, while being inside the circle indicates performance lower than the benchmark.}
 (b) Training dataset  \textbf{DR-VQA}. The attribute classification extracts disease phrases and related attributes (severity level and location) from the report. (c) Typical VQA examples of \textbf{instruction set}.}
    \label{fig:overview}
\end{figure}

To address these limitations on both technical and application aspects, we propose OmniFM-DR, a multi-modal model for reading chest radiographs, by providing more detailed evidence of associated diseases instead of rushing to the diagnosis directly. For the proposed multi-task transformer model, we unify the input and output labels of all sub-tasks into a uniform format for consistent modeling and joint training, which is detailed in the customized instruction tuning section. Figure \ref{fig:overview} illustrates the four main tasks performed by OmniFM-DR and provides an illustrative comparison of model capabilities in each task with other state-of-the-art (SOTA) methods, measured in each individual evaluation metric. OmniFM-DR is designed to handle a wide range of downstream tasks relevant to chest X-ray analysis, including diagnosis of common thoracic diseases and localization of those image-visible disease patterns. Additionally, it can perform image segmentation for pneumothorax, lungs, and heart regions. Most importantly, OmniFM-DR is capable of generating reports summarizing the findings by leveraging all the provided evidence mentioned here. Notably, OmniFM-DR is capable of reaching equivalent or even better performance compared to SOTA models (dedicated to specific tasks) in all downstream clinical applications, showing its effectiveness and generality in chest X-ray interpretation. 

Our contributions are three-fold:

% • We propose the first  model for multi-task analysis of Chest X-ray images, supporting four types of downstream tasks consisting of 8-class disease localization, report generation, 14-class classification, and 3-class segmentation.  The proposed model has achieved satisfactory performance of generalization and interpretability.
\begin{itemize}

\item 
Our proposed model offers a versatile approach to analyzing chest X-ray images, allowing for comprehensive and accurate radiography image analysis across various application tasks. It enhances the interpretability of chest X-ray reporting by generating more detailed information on disease attributes. This includes disease size, location, severity, and contour, providing stronger evidence for diagnosis and treatment.

\item We develop a unique framework for building datasets that are tailored for customized instruction tuning. Unlike the conventional method of organizing pair-wise supervision (consisting of an image and its corresponding label), our framework involves cross-task training supervision for each sample, which enhances the learning of correlations among tasks. We will release the composed datasets with detailed image-instruction-label triplets, which is the very first dataset of this kind to our knowledge.  

% \item The interpretability of the generated reports is enhanced by incorporating multiple tasks, enabling more transparent decision-making. Disease descriptions, such as disease position and severity, are inferred from the disease localization and segmentation tasks, further contributing to interpretability.

\item We applied the proposed model on various downstream application benchmarks, and an overall superior performance is shown compared to SOTA approaches. Furthermore, we conducted a controlled trial and evaluation on the generated reports performed by three radiologists. In the blinded comparison of 160 retrospective cases from four centers, three radiologists perceived the quality of \textcolor{black}{66\%} of the generated reports to be equivalent to or even better than the original physician reports. 
% Additionally, the generated reports have an average omission rate of 2.53\% across the four centers, which closely aligns with the radiologists' rate of 1.25\%.

% radiologists consistently preferred the OmniFM-DR generated reports over those originally from radiologists in approximately 60\% of the cases. Additionally, our suggested model demonstrates an average of 0.25 clinically significant errors per report.

% \item We also release the first comprehensive dataset, which includes radiograph reports, 14-class classification, 8-class disease localization, and corresponding lung, heart, and pneumothorax segmentation. Notably, the sub-dataset for pneumothorax segmentation is currently the largest available dataset to our knowledge.

\end{itemize}

% \begin{figure}[t]
% 	\centering
% 	%\fbox{\rule[-.5cm]{4cm}{4cm} \rule[-.5cm]{4cm}{0cm}}
%     \includegraphics[width=0.5\linewidth]{0906fig5.png}
% 	\caption{\textbf{DIVIDE INTO TWO, COMBINE WITH FIGURE 1 AND 4.} We introduce an extensive chest X-ray
% multi-task labeling dataset with uniform and fine-grain anatomical annotations for MIMIC-CXR databases. Our dataset contains heart\&lung contour (199257 cases) and pneumothorax contour (233 cases) for segmentation task, and disease phrase with related descriptor (135791 cases) for report instruction. The original images are sourced from the MIMIC-CXR dataset, which contains 24334 images and the accompanying 14 diseases classification, radiology reports, and bounding box of 8 diseases (1047 cases) for the detection task.}
% 	\label{fig:dataset}
% \end{figure}

\section{Results}

\subsection{Overview}
 
As illustrated in Figure \ref{fig:overview}, we train the multi-task model for analyzing chest X-ray images with a dataset specially designed for customized instruction tuning. It aims at a comprehensive analysis of chest X-ray images (providing detailed evidence) and enhanced interpretability (evidence-based diagnosis) as a tool for computer-aided diagnosis.
For the performance evaluation purpose, we applied the proposed model to various downstream tasks and benchmarks, and an overall superior performance is achieved compared to SOTA models (dedicated to each individual sub-task). Furthermore, we conducted a controlled trial for the evaluation of the generated report in comparison to the original reports, assessed by a group of radiologists. 
In a blinded comparison involving 160 historical reports from four different centers, three radiologists consistently rated the quality of the generated reports as comparable to or better than original radiological reports, with a success rate of over \textcolor{black}{66\%.} Moreover, our proposed model exhibited an omission rate of \textcolor{black}{1.87\% (95\% CI 1.06-2.83\%)} and an error rate of \textcolor{black}{2.24\% (95\% CI 1.39-3.20\%)} per report, which are close to those of the radiologist-provided reference reports (i.e., 1.25\% \textcolor{black}{(95\% CI 0.67-2.00\%)} and 2.00\% \textcolor{black}{(95\% CI 1.19-2.89\%)}). 

% \textbf{Downstream application datasets and benchmarks}
In Table~\ref{tab:MultiMedBench}, we illustrate the overall performance of the proposed OmniFM-DR across four main tasks on nine datasets, along with individual SOTA results, using a total of 140 thousand testing cases.
% labeled with 16 different radiographic classes. 全文中再没有提到16个类别
Our unified transformer model achieves superior results among the majority of the tasks on unseen datasets when performing direct inference. In the fine-tuning setting, the advantage remains significant in most of the tasks and metrics. 
Here, the proposed model is trained using a list of public datasets (detailed below). ChestXray14 and RSNA pneumonia datasets are utilized to evaluate the multi-label classification task performance, while MS-CXR, ChestXray14, and RSNA pneumonia datasets are used for the disease localization task. 
Furthermore, the largest available dataset for CXR reports (i.e., MIMIC-CXR) is utilized to evaluate the report generation performance. 
% Detailed comparations of the specific disease categories with other methods are provided in the supplements Table~\ref{tab:report}.
The metrics (AUC and F1)  refer to the macro average on the 14 diseases for ChestXray14. Accuracy (ACC) and mean Intersection over Union (mIoU) are utilized for the evaluation of disease localization, while Dice is utilized for segmentation tasks in three datasets (JSRT, CheXMark, and MS-PS). Furthermore, the performance of the report generation task is assessed by both clinical efficacy (CE, i.e., F1, Precision, Recall) and natural language processing (NLP, i.e., BL-4, METEOR, Rouge-L) metrics.

\begin{table}[bth!]
    \caption{\textcolor{black}{Performance comparison on MultiMedBench with the direct inference and fine-tune setting. We compare OmniFM-DR with specialist SOTA models. Across all tasks, datasets, and metrics combinations in MultiMedBench, we observe OmniFM-DR performance is equivalent to or exceeding SOTA. The results of OmniFM-DR are displayed with 95\% confidence intervals shown in parentheses.}}
    \centering
    \begin{tabular}{lllllll}
	\toprule
	\cmidrule(r){1-7}
           Task  &Dataset &Metric &\multicolumn{2}{c}{Direct Inference} & \multicolumn{2}{c}{Finetuning}\\
		           &        & &SOTA & OmniFM-DR &SOTA & OmniFM-DR \\
		\midrule   
   Classification  &ChestXray14 &AUC  &72.4\cite{wu2023medklip}   &\textbf{73.7} \scriptsize{(72.8, 74.6)} &77.8\cite{zhang2022contrastive} &\textbf{78.1} \scriptsize{(77.2, 79.0)} \\
                    &&F1&24.3\cite{wu2023medklip}&\textbf{26.4} \scriptsize{(25.0, 27.8)} &\textbf{33.0}\cite{zhang2022contrastive} &32.7 \scriptsize{(31.2, 34.2)} \\
                    &  RSNA   &AUC  &83.4\cite{Boecking_2022}   &\textbf{84.5} \scriptsize{(83.1, 85.9)}	        &88.6\cite{Huang2021GLoRIAAM}  &\textbf{89.1} \scriptsize{(88.2, 90.0)} \\
                    &  &F1	  &58.7\cite{zhang2022contrastive}  &\textbf{59.6} \scriptsize{(58.3, 60.9)} &\textbf{67.4}\cite{Huang2021GLoRIAAM} &66.5 \scriptsize{(64.5, 68.5)} \\  
 Localization      
                    &ChestXray14&ACC  &31.1\cite{Deng_2021_ICCV} &\textbf{56.3} \scriptsize{(55.7, 56.9)}& 50.7\cite{Zhu_2022}	&\textbf{60.9} \scriptsize{(59.7, 62.1)}\\
                    &           &mIoU &32.0\cite{Deng_2021_ICCV}	&\textbf{49.0} \scriptsize{(48.4, 49.6)}&46.9\cite{Zhu_2022}&\textbf{51.3} \scriptsize{(50.9, 51.7)}\\
                    & MS-CXR &ACC  &25.9\cite{Deng_2021_ICCV}&\textbf{46.7} \scriptsize{(44.9, 48.5)} & 45.8\cite{Zhu_2022}  &\textbf{55.4} \scriptsize{(53.7, 57.3)} \\
                    &           &mIoU &27.5\cite{Deng_2021_ICCV}	 &\textbf{46.4} \scriptsize{(46.1, 46.7)}	& 44.8\cite{Zhu_2022}  &\textbf{50.8} \scriptsize{(50.4, 51.2)} \\
                    &RSNA   &ACC &35.4\cite{Deng_2021_ICCV}&\textbf{42.5} \scriptsize{(41.7, 43.3)}	& 33.7\cite{Zhu_2022}	&\textbf{54.2} \scriptsize{(53.1, 55.3)} \\
                    &           &mIoU &37.1\cite{Deng_2021_ICCV}	 &\textbf{47.3} \scriptsize{(46.0, 48.6)}	& 33.6\cite{Zhu_2022}	&\textbf{49.7} \scriptsize{(49.4, 50.0)} \\
   Segmentation& JSRT &Dice &\textbf{93.6}\cite{Huang2021GLoRIAAM} &90.4 \scriptsize{(89.9, 90.9})&\textbf{95.0}\cite{Huang2021GLoRIAAM}&91.1 \scriptsize{(90.4, 91.8)} \\
                &CheXmask &Dice &\textbf{90.1}\cite{Huang2021GLoRIAAM}&88.6 \scriptsize{(88.2, 89.0)}&92.0\cite{Huang2021GLoRIAAM}&\textbf{93.3} \scriptsize{(92.6, 94.0)}\\
                &MS-PS   &Dice &48.7 \cite{Huang2021GLoRIAAM}&\textbf{59.6} \scriptsize{(57.5, 61.7)} &52.9\cite{Huang2021GLoRIAAM}&\textbf{66.2} \scriptsize{(64.9, 67.5)} \\
  Report generation & MIMIC-CXR &F1	    &30.6\cite{liu-etal-2021-contrastive} &\textbf{33.1} \scriptsize{(32.5, 33.7)} &-  &-  \\
                    &           &Precision&36.3\cite{liu-etal-2021-contrastive} &\textbf{42.8} \scriptsize{(40.0, 45.6)}&-  &- \\
                    &           &Recall   &30.1\cite{liu-etal-2021-contrastive} &\textbf{31.5} \scriptsize{(30.7, 32.3)}&-  &- \\
                    &      &RadGraph-F1   &17.2\cite{jeong2023multimodal} &\textbf{19.3} \scriptsize{(18.1, 20.5)}&-  &- \\
                    &           &BL-4     &\textbf{12.6}\cite{tanida2023interactive} &11.1 \scriptsize{(10.8, 11.4)}&-  &- \\
                    &           &METEOR   &\textbf{16.8}\cite{tanida2023interactive} & 14.1 \scriptsize{(13.9, 14.3)}&-  &- \\
                    &           &Rouge-L  &\textbf{28.4}\cite{liu-etal-2021-contrastive} & 26.5 \scriptsize{(26.2, 26.8)}&-  &- \\
		\bottomrule
    \end{tabular}
    \label{tab:MultiMedBench}
\end{table}

\subsection{Disease Classification}
\begin{figure}[t]
	\centering
	%\fbox{\rule[-.5cm]{4cm}{4cm} \rule[-.5cm]{4cm}{0cm}}
    \includegraphics[width=1.0\linewidth]{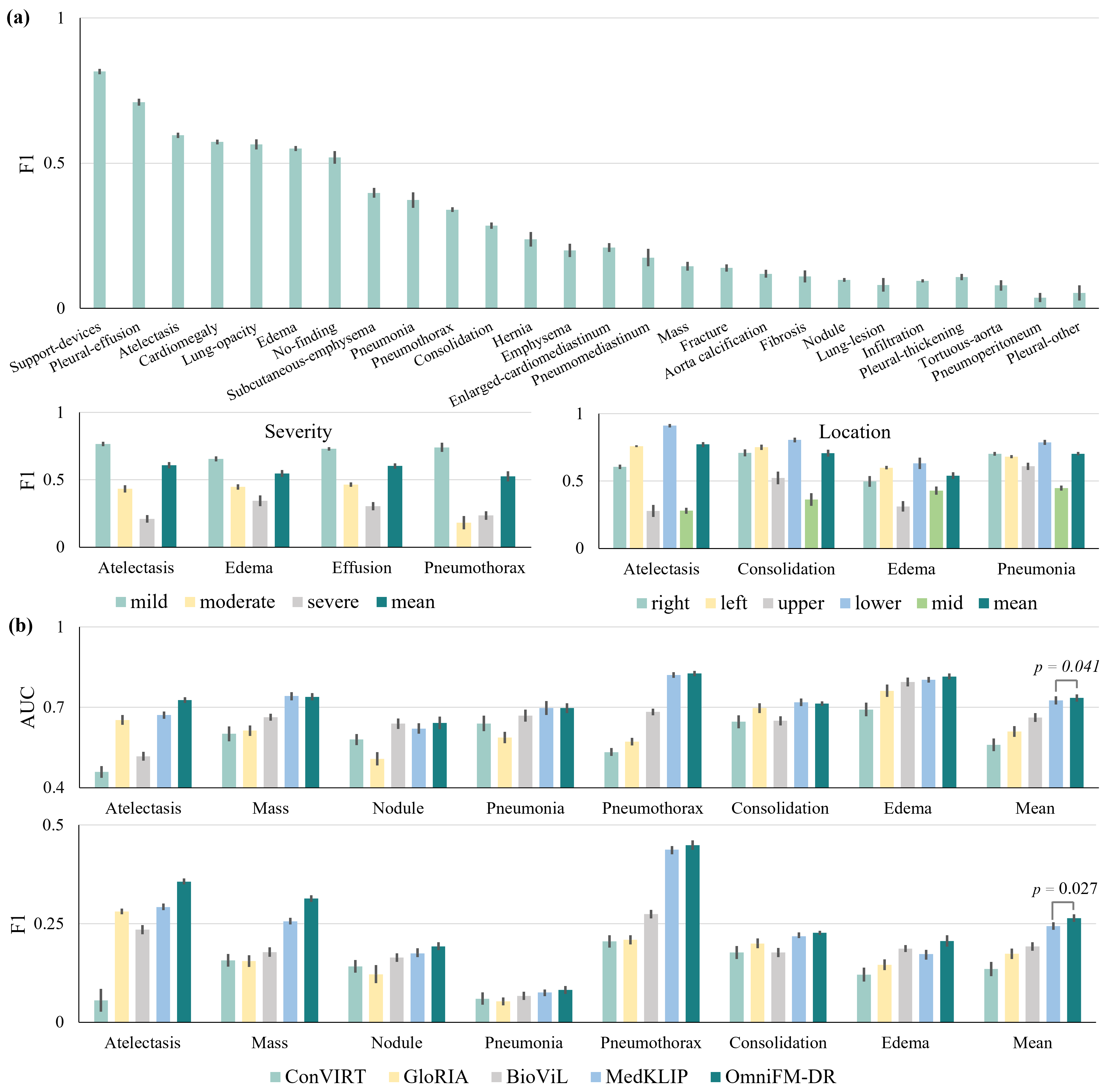}
	\caption{(a) In domain evaluation of 26 disease entities on the MIMIC-CXR dataset (upper panel) and attribute classification task in disease severity level and location (lower panel). The mean represents the micron average. (b) Out-of-domain evaluation between OmniFM-DR and other classification models (i.e., ConVIRT, GloRIA, BioViL, and MedKLIP) on the ChestXray14 dataset. AUC score and F1 are utilized to assess the classification task. The mean represents macro average. \textcolor{black}{The error bars of our method show 95\% confidence intervals, and the bars’ center represents the mean value of the AUC or F1.}}
	\label{fig:cls}
\end{figure}

We explore two types of classification tasks: disease entity classification and attribute classification. The entity classification task focuses on classifying disease categories, while the attribute classification task determines the disease attributes, e.g., location and severity. 
% corresponding to instructions (i.e. "What disease does the image have?",  "Is atelectasis in this image?"). On the other hand, the attribute classification task determines the location and severity level of the disease with instructions (i.e. "Where is pneumothorax?", "what is the severity of pneumothorax?").  
As illustrated in Table~\ref{tab:MultiMedBench}, we evaluate the disease entity classification task on the ChestXray14 and RSNA Pneumonia datasets with both direct inference and fine-tuning settings. 
Compared to the previous best results, Our model shows improvements of \textcolor{black}{1.3\%} in AUC and \textcolor{black}{2.1\%} in F1 for the ChestXRay14 dataset, with improvements of \textcolor{black}{1.1\%} in AUC and \textcolor{black}{0.9\%} in F1 for the RSNA Pneumonia dataset with direct inference setting, in comparison to models explicitly trained for each classification task. We further conduct the fine-tuning experiments and notice the average results of various methods are similar when models are fine-tuned with 50 samples for each finding on the target dataset. 
% Our model shows improvement in the entity classification task assessed by AUC and F1 metrics. 
Figure\ref{fig:cls}(a) further shows the detailed distribution of F1 results for 26 different disease entities that are aligned with the long-tail distribution on the MIMIC-CXR dataset.
Additionally, our model attains high F1 scores across the diseases in the attribute classification task. For instance, the severity classification ACC for Effusion and Pneumothorax reaches \textcolor{black}{60.3\% (95\% CI 59.1-61.5\%)} and \textcolor{black}{52.6\% (95\% CI 50.8-54.4\%)}, respectively.
% with an average of 59.2\% for seven diseases with attribute descriptions.
% In addition, localization classification achieves an ACC of 63.2\% for Effusion and 63.6\% for Pneumothorax. On average, the ACC and F1 for localization classification across eight diseases reach 53.3\% and 50.3\%, respectively.
% (Supplementary Table~\ref{tab:cls}). 
% \colorbox{yellow}{[NTU]}  

% The 95\% confidence interval is shown in (a) and (b)}
When evaluating the domain-shifted ChestXray14 dataset, our method achieves satisfactory results comparable to SOTA results on ten diseases (i.e., Atelectasis, Mass, Nodule, Pneumonia, Pneumothorax, Consolidation, Edema, etc.) shown in Figure \ref{fig:cls}(b).
For instance, our method achieves an AUC and F1 of \textcolor{black}{72.8\% (95\% CI 72.1-73.5\%) and 35.7\% (95\% CI 34.6-36.8\%)} for Atelectasis and \textcolor{black}{64.2\% (95\% CI 63.2-65.2\%) and 19.3\% (95\% CI 18.1-20.5\%)} for Nodule, respectively (see
Supplementary Table~\ref{tab:cls}).
These results surpass the previous best results, demonstrating the effectiveness of our method in accurately identifying and localizing relatively small abnormalities in Chest X-ray images.  
\textcolor{black}{Additionally, the performance improvement of the proposed method on the disease classification is statistically significant compared to the SOTA ( with \textit{p} < 0.05).}
 % Quantitative details are available in Supplementary Table~\ref{tab:cls}.
 
\subsection{Disease Localization}
We herein conduct extensive experiments on ChestXray14, MS-CXR, and RSNA Pneumonia datasets to evaluate disease localization under both direct inference and 20-shot fine-tuning settings. 
Under the direct inference setting, our model generally achieves the best performances over all existing visual grounding models. As shown in Table~\ref{tab:MultiMedBench}, OmniFM-DR gets an ACC of \textcolor{black}{56.3\% (95\% CI 55.7-56.9\%), 46.7\% (95\% CI 44.9-48.5\%), and 42.5\% (95\% CI 41.7-43.3\%)} on the ChestXray14, MS-CXR, and RSNA Pneumonia datasets respectively, surpassing other methods by a large margin, 7\% to 25\%.
When fine-tuned on the downstream dataset with 20 samples for each label, OmniFM-DR consistently scores the highest ACC of \textcolor{black}{60.9\% (95\% CI 59.7-62.1\%), 55.4\% (95\% CI 53.7-57.3\%), and 54.2\% (95\% CI 53.1-55.3\%)} on the three datasets. 
% For full-data fine-tune, OmniFM-DR consistenly achieves satifactory performance, surpasses previous soft by 19.0\%, 21.4\%, 29.2\% on three test datasets respectively.
% Note that the other disease localization methods (i.e., VGTR, SeqTR, TransVG) are trained on the RefCoco dataset and have inferior direct inference ability on the Xray data and perform better after fine-tuning. 

\begin{figure}[t]
	\centering
	%\fbox{\rule[-.5cm]{4cm}{4cm} \rule[-.5cm]{4cm}{0cm}}
    \includegraphics[width=1\linewidth]{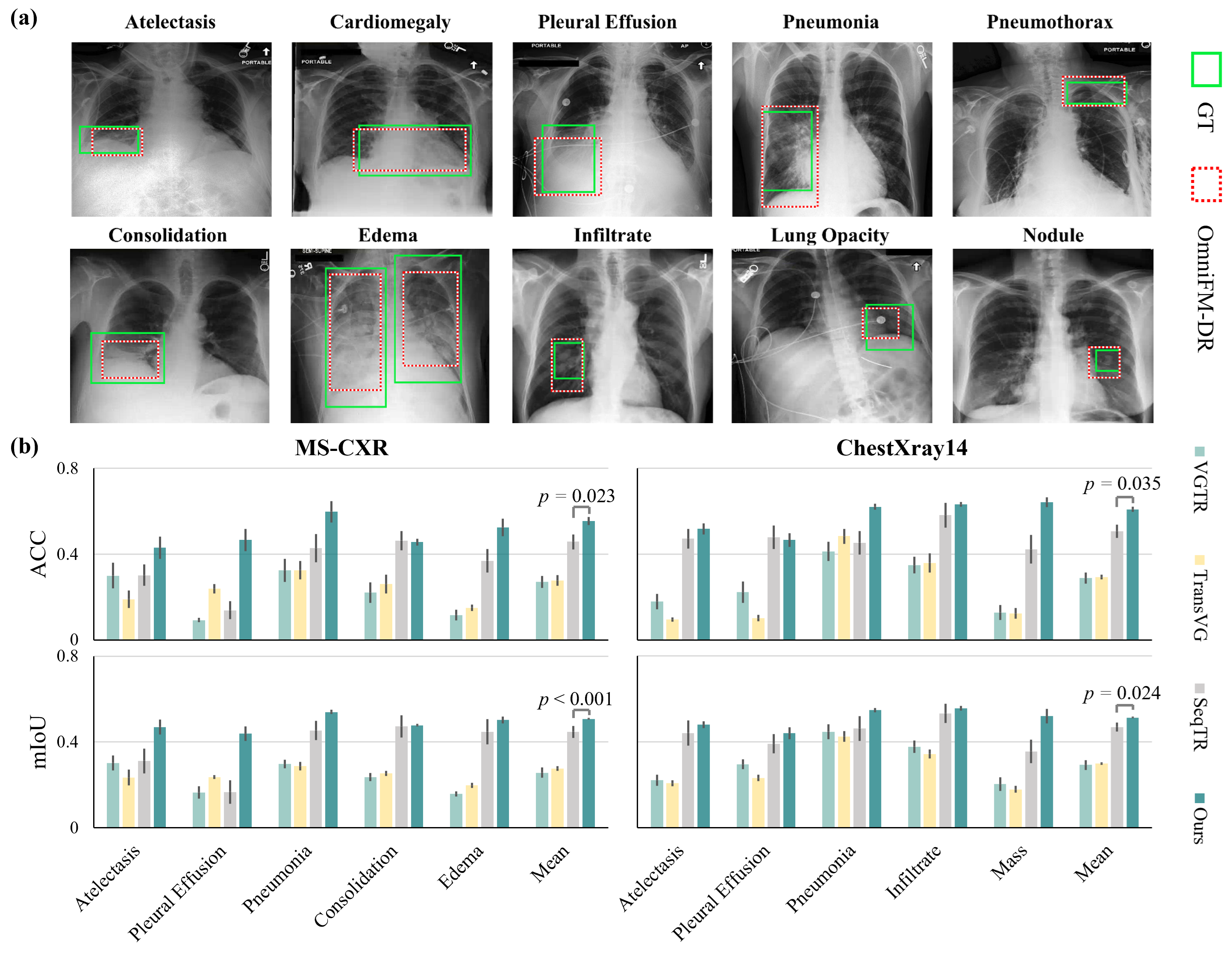}
	\caption{Assessment of the disease localization task with ACC and mIoU metrics on the MS-CXR and ChestXray14 datasets. (a) Examples with bounding box prediction and (b) comparisons between OmniFM-DR and other methods. \textcolor{black}{The error bars show 95\% confidence intervals, and the bars’ center represents the mean value of the ACC or mIoU.}
 % The two datasets share five common diseases: Cardiomegaly, Effusion, Pneumothorax, Atelectasis, and Pneumonia. Additionally, the MS-CXR dataset includes three additional diseases: Consolidation, Edema, and Opacity, while the ChestXray14 dataset includes three additional diseases: Infiltrate, Mass, and Nodule. 
 }
	\label{fig:vg}
\end{figure}

 Figure \ref{fig:vg} further presents the model's capability of disease localization across multiple diseases and comparations of OmniFM-DR against three approaches (i.e., VGTR, TransVG, SeqTR) on MS-CXR and ChestXray14 dataset with fine-tuning setting. The two datasets share five common diseases: Cardiomegaly, Effusion, Pneumothorax, Atelectasis, and Pneumonia. 
 Moreover, the MS-CXR dataset comprises three additional diseases (i.e., Consolidation, Edema, and Opacity), while the ChestXray14 dataset includes Infiltrate, Mass, and Nodule.
 It is observed that OmniFM-DR is capable of accurately identifying disease locations across multiple diseases.
 % (i.e., Opacity, Effusion, Pneumothorax, Pneumonia, Edema, Atelectasis, and Cardiomegaly). 
 Furthermore, the model consistently achieves large advantages over other approaches for most disease categories. For seven of eight diseases on the MS-CXR dataset, OmniFM-DR gets significantly higher ACC and mIoU than the best-performing baseline,  with the largest improvements in Pleural Effusion (>22\%).
 Supplementary Table~\ref{tab:VG} further provides detailed comparisons of the eight diseases among OmniFM-DR and other approaches. 
 % Our method outperforms the others in terms of accuracy and mIoU for various diseases.
% We further test the model’s performance by varying the number of images.
% When 100-shot data is used for training, 
% OmniFM-DR exhibits large performance improvements with respect to these baselines. 
 % For instance, OmniFM-DR 

\subsection{Segmentation}
% In our multi-task approach, 
We adopt a polygon-based contour representation for the segmentation task to achieve a uniform input-output format with other tasks, i.e., predicting a list of polygon vertexes instead of region masks. As shown in Table~\ref{tab:MultiMedBench}, the Dice coefficient is utilized for evaluating the segmentation of lung and heart contours. 
% Our unified model is comparable to pixel-based segmentation methods. 
\textcolor{black}{For the segmentation of the heart and lung contours, both our unified model and the pixel-based segmentation method achieved Dice scores around 90\%. However, for the segmentation of pneumothorax, our method achieved better results with a significant improvement of 10\% in Dice score, both in direct inference and finetuning settings.}

\begin{figure}[t]
	\centering
	%\fbox{\rule[-.5cm]{4cm}{4cm} \rule[-.5cm]{4cm}{0cm}}
    \includegraphics[width=1\linewidth]{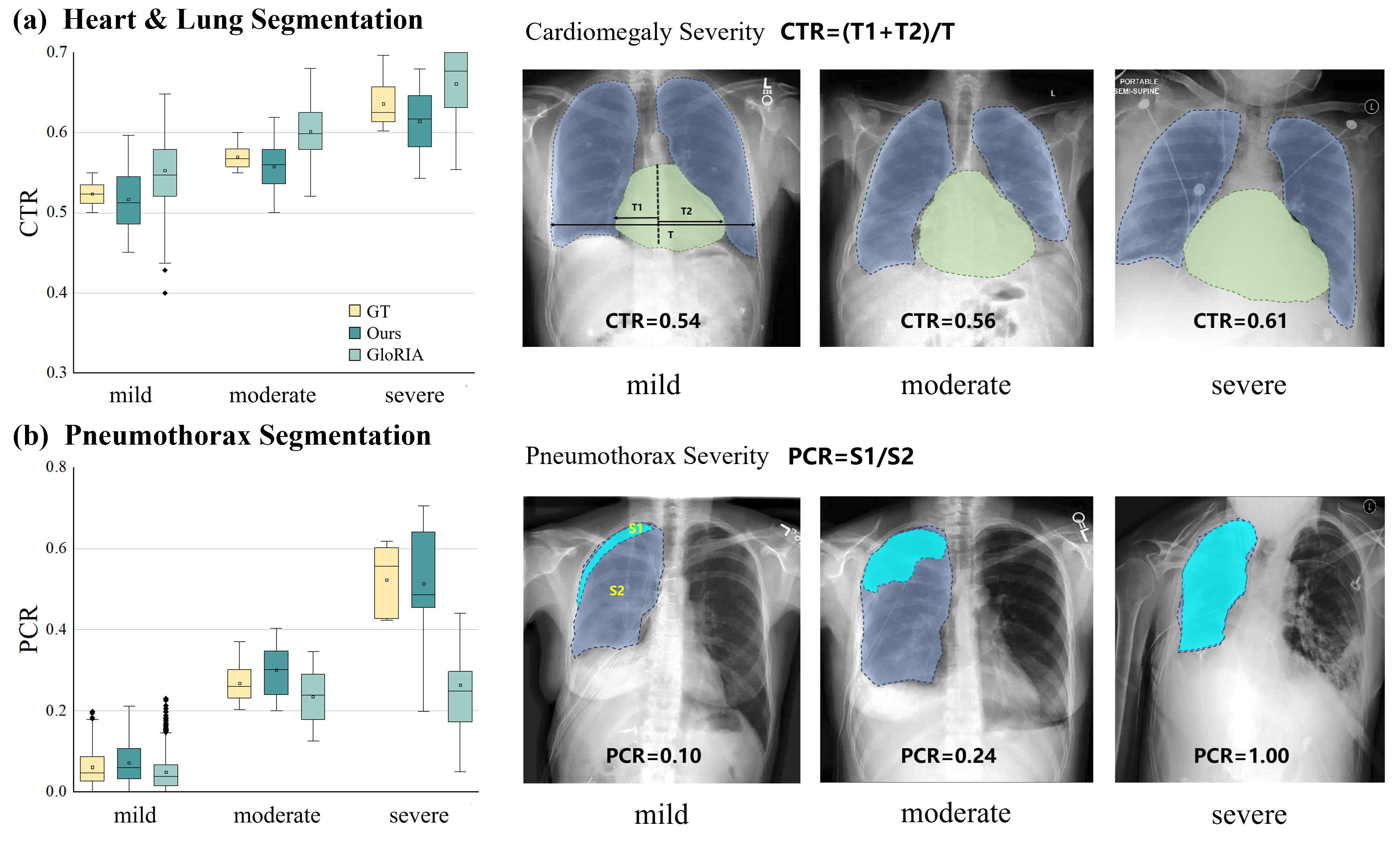}
	\caption{
 % Results of heart, lung, and pneumothorax segmentation. 
 % % OmniFM-DR represents the segmentation mask by outputting a series of coordinate points along the contour. 
 % (a) Segmentation of the heart and lung can be used to assess the severity of cardiomegaly. 
 % % The cardiothoracic ratio (CTR), calculated as the ratio between the maximum transverse diameter of the heart and the chest, is used to evaluate the cardiomegaly severity as mild, moderate, or severe. 
 %    (b) Pneumothorax segmentation is performed to assess the severity of pneumothorax. 
 %    % The area method is one of several techniques used to calculate the pneumothorax compression ratio (PCR), which is represented by the ratio of the pneumothorax area to the affected lung area.
 \textcolor{black}{OmniFM-DR's performance in heart, lung, and pneumothorax segmentation. (a) Segmentation of the heart and lung can be used to assess the severity of cardiomegaly by calculating the CTR value. (b) Pneumothorax segmentation is performed to assess the severity of pneumothorax by calculating the PCR value. The left box chart in (a) and (b) shows the comparison between the predicted results of CTR/PCR by our model and GloRIA, as well as the distribution of the GT under different levels of severity for cardiomegaly/pneumothorax.}}
	\label{fig:segmentation}
\end{figure}

To assess the severity of cardiomegaly and pneumothorax in potentially more detailed analyses, we perform post-processing on the segmented lung and heart masks to calculate the cardiothoracic ratio (CTR) and pneumothorax ratio (PCR). CTR is calculated as the ratio between the maximum transverse diameter of the heart and the chest, commonly used to evaluate the cardiomegaly severity as mild, moderate, or severe. The area method is used to calculate the PCR, which is represented by the ratio of the pneumothorax area to the affected lung area. Figure \ref{fig:segmentation} illustrates the segmentation results of cardiomegaly and pneumothorax with different severity on the test splits of the CheXmask dataset and the SIIM dataset. \textcolor{black}{We observe that the overall performance of lung and heart contours is satisfactory across different CTR (see Figure 4(a) box plot). GT achieves mean values of 0.52/0.57/0.63 on mild/moderate/severe CTR, while our model achieves 0.51/0.56/0.62 and GloRIA achieves 0.55/0.60/0.68. Furthermore, pneumothorax exhibits significant variations in location and size, and our model exhibits higher accuracy in predicting severe pneumothorax compared to GloRIA (see Figure 4(b) box plot). GT achieves mean values of 0.05/0.26/0.56 on mild/moderate/severe PCR, while our model achieves 0.06/0.30/0.49 and GloRIA achieves 0.04/0.24/0.25.} For more accurate segmentation of mild pneumothorax, we incorporate a segmentation head with a U-Net decoder and achieve a Dice value of 59.6\% \textcolor{black}{(95\% CI 57.5-61.7\%)} and 66.2\% \textcolor{black}{(95\% CI 64.9-67.5\%)} under direct inference and fine-tune settings.

\subsection{Report Generation}
\begin{figure}[t]
	\centering
	%\fbox{\rule[-.5cm]{4cm}{4cm} \rule[-.5cm]{4cm}{0cm}}
    \includegraphics[width=1\linewidth]{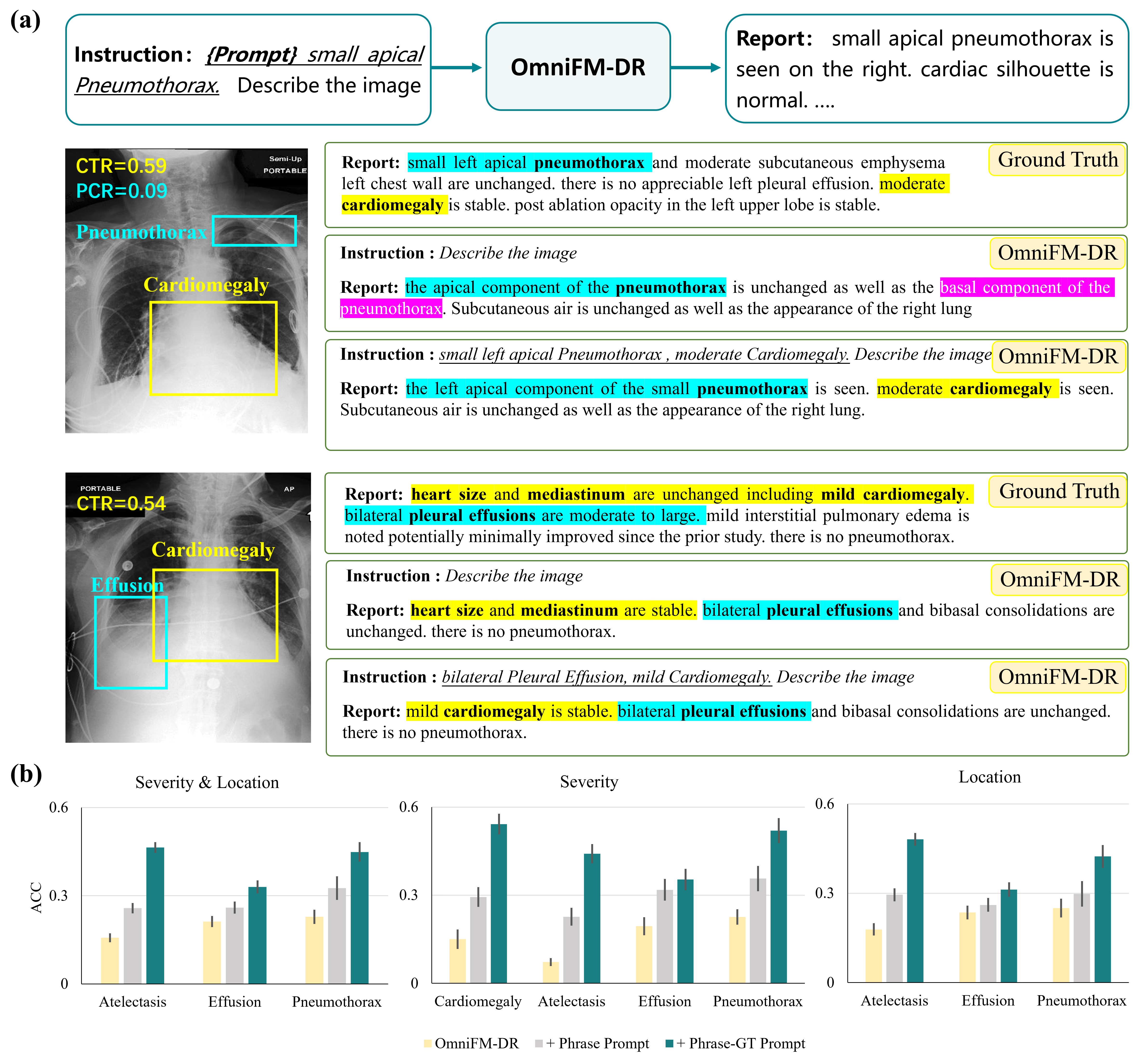}
	\caption{(a) Comparison results of two examples (e.g., Pneumothorax and Pleural Effusion). We further compare the generated report with and without designed instruction. The customized instruction includes the disease category, location, and severity level that are inferred from other tasks.
    (b) Comparison of generated reports in three ways: without prompt, with prompt from multi-task results, and with prompt from ground truth (GT). We use ACC as the evaluation metric. \textcolor{black}{The error bars show 95\% confidence intervals, and the bars’ center represents the mean value of the ACC.
    % The 95\% confidence interval is shown in (b)}.
    }}
	\label{fig:prompt}
\end{figure}

As the summary of the radiology reading process, reports contain major findings and possible disease diagnoses from the radiologists. As the unique feature of our proposed framework, we hypothesize adding pre-generated evidence could significantly improve the quality of AI-generated reports. Figure \ref{fig:prompt} demonstrates this quality improvement with the disease attributes prompt. 
During the inference stage, the customized prompt is initially derived from the classification task, including disease entity, severity, and rough location. The disease attribute prompt will be updated when a more accurate bounding box or severity of Pneumothorax/Cardiomegaly is available from the disease localization or segmentation task.  
The generated reports of typical cases of Pneumothorax and Pleural Effusion are provided in Figure~\ref{fig:prompt}(a). The proposed model is capable of identifying the diseases with an accurate disease localization box and generated report. 
Detailed descriptions of pneumothorax entities and attributes such as "small left apical pneumothorax" are highlighted by blue text in the ground truth report and are well predicted in the generated report. The location and severity of the pneumothorax are further verified by the bounding boxes and calculated metric (i.e., PCR). On the other hand, the descriptions of disease attributes (e.g., "small" of pneumothorax, "moderate cardiomegaly") are omitted in the generated report without proper prompt. 
For the Effusion case, we also find the importance of proper input prompts in the task of report generation.
The descriptions of pleural effusion and cardiomegaly are more detailed and accurate in the generated report with prompt. The bounding box of pleural effusion and mild cardiomegaly is indicated by the disease localization and segmentation task (i.e., CTR=0.54). With more specific prompts, the generated report shows accurate descriptions as "bilateral pleural effusion, no pneumothorax, and mild heart size."
In contrast, the general model without the designed prompt provides an inaccurate assessment of cardiomegaly, as described as "stable heart size."  

Figure \ref{fig:prompt}(b) further compares the accuracy of three different conditions (i.e., baseline, with phrase prompt, and with phrase-GT prompt).
With the help of disease attributes prompt, the accuracy of severity and location description has been improved \textcolor{black}{by 10.1\%, 4.8\%, 9.8\% for Atelectasis, Effusion, and Pneumothorax, respectively. The accuracy of Cardiomegaly in severity description is improved by 14.4\% (Cardiomegaly does not have a location attribute).} When the ground truth disease phrase is utilized as the prompt, we find the quality of the report improved further as an upbound for our model.

\textbf{Subjective Comparison Study}
% \subsection{Radiologist evaluation}
To assess the clinical interpretation, radiologists' evaluations are employed to examine the quality of radiology reports. \textcolor{black}{We have engaged three experienced radiologists with ten, six, and five years of professional practice, respectively, to participate in this study.}
The side-by-side comparison study focuses on errors associated with the presence, location, and severity of clinical findings. Non-clinical errors, such as referring to views or previous studies that do not exist, are excluded from our evaluation.
% In parallel evaluation, 

\begin{figure}[t]
	\centering
    \includegraphics[width=1\linewidth]{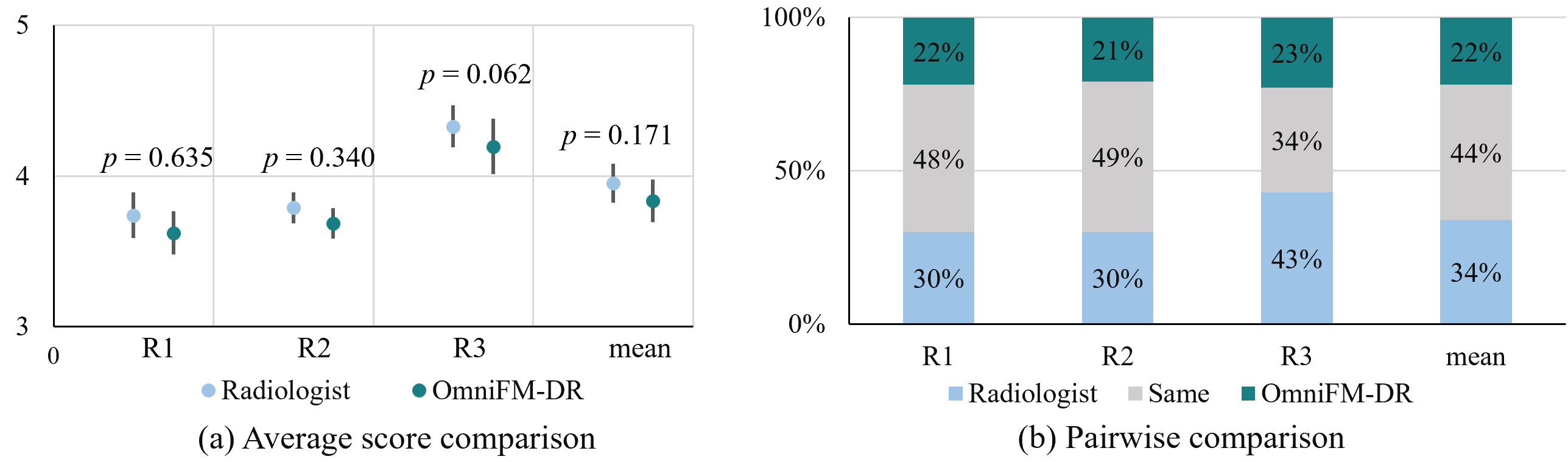}
	\caption{\textbf{Side-by-side comparison}. Three radiologists reviewed and scored the clinically derived reports from four centers and reports generated by OmniFM-DR. 
 \textcolor{black}{
 The error bars show 95\% confidence intervals, and the bars’ center represents the mean value of the average scores.
 % The 95\% confidence interval is shown in (a)}.
 }}
	\label{fig:sbs-evaluation}
\end{figure}

% In the side-by-side comparison,
% Figure \ref{fig:evaluation}(a) presents the average score comparison result by three radiologists. Three radiologists provided average results for the original reports and generated reports in four centers: 4.01/3.61, 4.03/3.84, 4.08/3.63, and 3.76/3.68. The average result across the four centers is 3.97/3.69, indicating that the quality of the generated reports is comparable to the original doctor reports. Figure \ref{fig:evaluation}(b) presents the pairwise comparison result, radiologists believe that 54\%/64\%/54\%/62\% of the generated reports are of equal or better quality than the original doctor reports at four centers, respectively. Despite the observed fluctuations in results across different centers, the collective analysis suggests that our generated reports exhibit a comparable level of quality to the original reports generated by medical professionals.
Figure \ref{fig:sbs-evaluation}(a) presents the average score comparison result from three radiologists. The average ratings for the radiologist-provided reference reports and generated reports are 3.74 \textcolor{black}{(95\% CI 3.62-3.86) / 3.62 (95\% CI 3.48-3.76)}, 3.79 \textcolor{black}{(95\% CI 3.69-3.89) / 3.69 (95\% CI 3.59-3.79)}, and 4.33 \textcolor{black}{(95\% CI 4.19-4.47) / 4.20 (95\% CI 4.02-4.38)}, respectively. The average result across three radiologists is 3.95 \textcolor{black}{(95\% CI 3.82-4.08) / 3.84 (95\% CI 3.74-3.94)}. \textcolor{black}{We utilize the Chi-Square test to verify the significance of performance differences between the model-generated reports and the original ones across different data centers. 
% The p value of radiologists 1, 2, and 3 are 0.645. 0.340, 0.031, respectively. 
With the computed \textit{p} value of three tests (compared to the original reports by three doctors) are 0.635, 0.340, and 0.062, respectively, it indicates that the generated reports are comparable to the reference reports in overall quality.} Figure \ref{fig:sbs-evaluation}(b) presents the pairwise comparison result. Three radiologists believe that the quality of generated reports is equal to or even better than the original reports by \textcolor{black}{70\%, 70\%, and 57\%}, respectively. Despite the observed fluctuations in results across radiologists, the collective analysis suggests that our generated reports exhibit a comparable level of quality to the original reports generated by medical professionals.

\begin{figure}[t]
	\centering
    \includegraphics[width=1\linewidth]{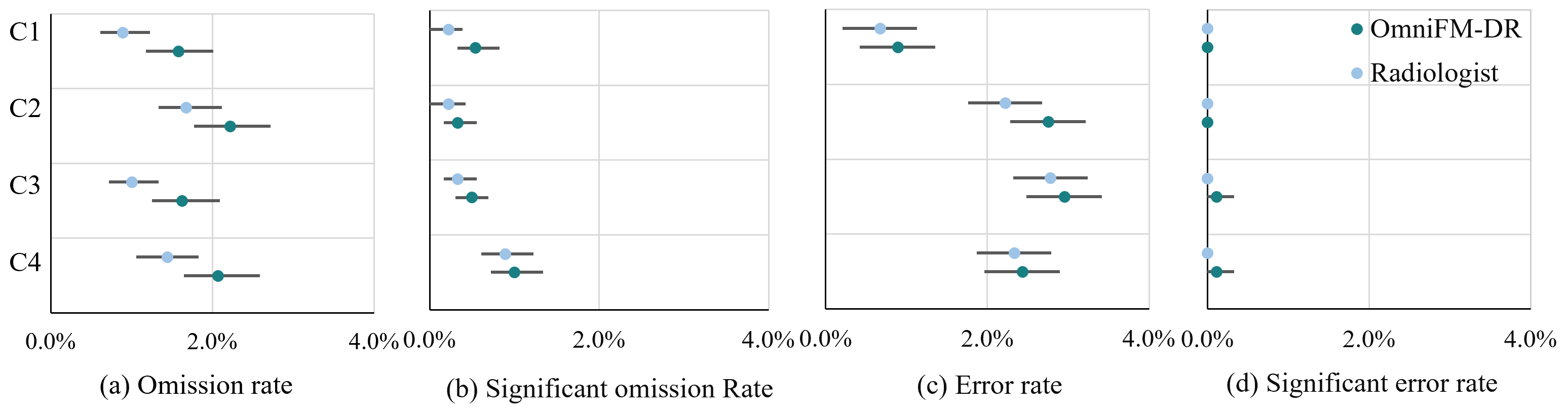}
	\caption{\textbf{Independent evaluation} of omissions and clinical errors for reports generated by OmniFM-DR. Significant errors are related to the presence, location, or severity of clinical findings, which are identified by radiologists. \textcolor{black}{
 The error bars show 95\% confidence intervals, and the bars’ center represents the mean value.}}
	\label{fig:ie-evaluation}
\end{figure}

In the detailed evaluation, \textcolor{black}{we show results of omission rate and error rate.} Omission rate refers to missed disease diagnoses, and error rate refers to inaccuracies in severity or location descriptions or false disease diagnosis. We report the results on the report level. Figure \ref{fig:ie-evaluation}(a) shows the total omission rate for the reference and generated reports in the four centers are as follows: 0.89\% \textcolor{black}{(95\% CI 0.33-1.56\%) / 1.58\% (95\% CI 0.78-2.44\%)}, 1.67\% \textcolor{black}{(95\% CI 1.00-2.56\%) / 2.22\% (95\% CI 1.33-3.22\%)}, 1.00\% \textcolor{black}{(95\% CI 0.44-1.67\%) / 1.62\% (95\% CI 0.89-2.56\%)}, 1.44\% \textcolor{black}{(95\% CI 0.67-2.22\%) / 2.07\% (95\% CI 1.22-3.11\%)}. 
% On average, the omission rate for the generated reports 2.53\% is close to that of the radiologist-provided reference reports, i.e., 1.25\%, with a significant omission rate of 0.42\%/1.00\%. 
Figure \ref{fig:ie-evaluation}(c) shows the error rate, i.e., inaccuracies in severity or location descriptions or false positives. The total error rate for the reference and generated reports in the four centers are 0.67\% \textcolor{black}{(95\% CI 0.22-1.22\%) / 0.89\% (95\% CI 0.33-1.56\%)}, 2.22\% \textcolor{black}{(95\% CI 1.33-3.22\%) / 2.75\% (95\% CI 1.89-3.89\%)}, 2.78\% \textcolor{black}{ (95\% CI 1.78-3.89\%) / 2.95\% (95\% CI 1.89-4.11\%)}, 2.33\% \textcolor{black}{ (95\% CI 1.44-3.22\%) / 2.36\% (95\% CI 1.45-3.24\%)}. \textcolor{black}{Based on the results, the omission rate and error rate observed in OmniFM-DR reports are comparable to those original ones from three data centers.}

% \subsection{A full-label dataset for chest x-ray images }
% Based on the MS-CXR dataset consisting of 1047 images, we supplemented it with segmentation results for pneumothorax, left lung, right lung, and heart. This enabled us to construct a full-label dataset for classification, disease localization, and segmentation tasks. The dataset demonstrates high Dice Similarity Coefficient (DSC) values ranging from 0.913 to 0.941, indicating a strong agreement between the GT and predicted masks. Additionally, the dataset underwent quality assurance by radiologists to ensure its accuracy and reliability.

% \subsection{Ablations}  
\section{Discussion}
% \subsection{ OmniFM-DR could handle multi-task analysis of Chest X-ray images and provides reliable evidence for the generated report} 
% \textbf{OmniFM-DR handles multi-task analysis of Chest X-ray images} 

Our research focuses on developing a single and unified model for Chest X-ray image analysis, with the goal of addressing multi-tasks for better clinical interpretation,  including disease classification, localization, segmentation, and report generation. 
By leveraging a carefully designed instruct tuning framework and incorporating diverse training strategies, our method can effectively extract relevant features and make predictions for multiple tasks in Chest X-ray images. This allows for a comprehensive analysis and diagnosis of various diseases and abnormalities in the images.
With unseen evaluation datasets, the proposed unified model has exhibited impressive performance in disease classification and localization tasks (via direct inference), surpassing the capabilities of existing state-of-the-art methods with a few-shot fine-tune setting. 

% \textbf{ OmniFM-DR provides reliable evidence for a better explainability} 
\textcolor{black}{OmniFM-DR provides reliable evidence for a better explainability.} 
It is crucial that automated medical report generators produce trustworthy, easily understandable, and accurate reports for effective utilization in practice. To achieve the desired goals, it is crucial to have high-quality explanations about how the report was generated and the process involved in reaching the diagnosis. An explainable system helps developers to pinpoint any weaknesses or inefficiencies, while clinicians can rely on the decisions made with the aid of these systems.

Here, we believe the interpretability of the reports can be validated mutually with the results obtained from other tasks within the model. For instance, the disease category, severity level, and approximate location of the lesion could be primarily verified with the disease entity classification and attribute classification task. The disease localization task could further provide a more accurate bounding box of the lesion. For Pneumothorax and Cardiomegaly, the segmentation function could provide an accurate assessment of disease degree by postprocessing the contours of the pneumothorax/lung/heart mask.
These results together contribute to a better verification of the generated reports.
 
Furthermore, we improve the explainability of the generated reports by incorporating customized prompts that provide information regarding disease attributes. 
Our results indicate that disease-specific prompts have improved recall and F1 score of disease entities — key metrics in evaluating the performance of automated report generation models. The results of this study underscore the significance of incorporating prompts (evidence) into the model, thereby contributing to the model's performance in the report generation process. This technique can prove to be a valuable strategy for improving the overall accuracy of AI models, specifically in the medical field. 

% \textbf{Image-instruction-label triplet dataset is designed for the promoting multi-task learning} 
\textcolor{black}{Image-instruction-label triplet dataset is designed for promoting multi-task learning.} 
While there exist various single-task datasets, there have been limited attempts to unify them and create benchmarks for the development of a single and more comprehensive model. As one of our major contributions, we design and plan to release a comprehensive dataset of Chest X-ray data. This dataset includes full-label annotations, enabling researchers and practitioners to explore and leverage the benefits of multi-task learning in this domain. By sharing this dataset, we aim to encourage and support further advancements in multi-task learning approaches for Chest X-ray analysis. This can potentially create new opportunities in clinical applications \cite{johnson2019mimic,irvin2019chexpert,wang2017chestx}. 
The benchmark has several limitations, including the limited size of the individual datasets and limited modality and task diversity. Another key barrier to developing models for use across an even wider variety of biomedical data types is the lack of large-scale multimodal datasets, which would permit joint learning and alignment of the modality-specific encoders with the decoder.

% \textbf{OmniFM-DR achieves comparable reporting results with expert radiologists} 
\textcolor{black}{OmniFM-DR achieves comparable reporting results with expert radiologists.} 
% The proposed model achieves satisfactory performance on all tasks. However, its out-of-the-box performance on MultiMedBench was poor and OmniFM-DR outperforms it by a wide margin across model scales. This result suggests that finetune with domain-specific biomedical data is
% critical to achieving good performance on biomedical tasks, perhaps due to the distribution shift presented by the domain overall compared to the plethora of non-medical tasks and modalities.
We conduct the evaluation of report quality based on the assessment of radiologists, which indicates that the model performs well on the complex multimodal task of generating radiology reports. In \textcolor{black}{66\%} of the cases, the generated reports are of equal or even better quality compared with the reference reports generated by medical professionals. Additionally, the average omission rate and error rate in the model-generated reports are similar to those found in reports generated by medical professionals on the same dataset. These promising findings demonstrate the rapid progress in the field of automatic radiology report generation and suggest the potential for future clinical applications.

% \textbf{Limitation and Future Work}
Our proposed model exhibits competent performance across multiple tasks and enhances the explainability of the generated results. However, it still faces challenges when it comes to generalizing to unseen disease categories and undefined instructions (tasks). These limitations highlight the need for further research and development to improve the model's ability to handle novel diseases and adapt to unfamiliar instructions. 
In the future, given the wider array of modalities and tasks in instruction tuning, more generalized models are expected to understand the tasks better and render modality-wise and task-wise emergent capability. 
\textcolor{black}{
Additionally, we have observed inconsistencies in the predictions of our model across different tasks. As part of our future plans, we aim to enhance consistency by improving the training data and refining the model. By addressing these issues, we expect to achieve better alignment and coherence in the predictions across various tasks. This may involve augmenting the training dataset to include more diverse examples, implementing regularization techniques, or fine-tuning the model architecture to encourage consistent predictions. These efforts will contribute to improving the overall performance and reliability of our model.}

% It is crucial that the encoders for such diverse modalities are incorporated into the joint training.

% \section{Conclusion}
% Our research focuses on developing a unified model for Chest X-ray data, with the goal of addressing multiple tasks within this domain using a single unified model. These tasks include report generation, classification, disease detection, and segmentation. The proposed unified model has exhibited impressive performance in direct inference classification and grounding tasks, surpassing the capabilities of existing state-of-the-art methods when fine-tuned. 
% One notable aspect of our model is the improved explainability of the generated reports. We achieve this by incorporating customized prompts that provide information regarding disease position and severity. This enhancement contributes to a better understanding of the reasoning behind the generated reports. The interpretability of the reports can be further validated by considering the results obtained from other tasks within the model.
% Additionally, as part of our contribution, we have released a comprehensive dataset of Chest X-ray data. This dataset includes full-label annotations, enabling researchers and practitioners to explore and leverage the benefits of multi-task learning in this domain. By sharing this dataset, we aim to encourage and support further advancements in multi-task learning approaches for Chest X-ray analysis.

\section{Material and Method}
%\label{sec:others}
\subsection{Building Dataset for Customized Instruction Tuning}
% Our curation of MultiMedBench is a step towards addressing this unmet need. However, the benchmark has suffered from the limited size of the individual datasets, and limited modality and task diversity. 
% Another key barrier to developing models for use across an even wider variety of biomedical data types is the lack of large-scale multimodal datasets, which would permit joint learning and alignment of the modality-specific encoders with the decoder.
% \subsubsection{DR-VQA Dataset Construction}
% AI progress to date has largely been catalyzed by the development of high-quality datasets. The availability of large public chest X-ray datasets recently has significantly contributed to the research community\cite{johnson2019mimic,irvin2019chexpert,wang2017chestx}. While there exist various single-task datasets, there have been limited attempts to unify them and create benchmarks for the development of a multi-task model. 
In this work, we construct a multi-task dataset for joint training of disease classification, localization, segmentation, and report generation. In general, we unify the input and output labels of all sub-tasks into a uniform format for consistent modeling and joint training, i.e., a set of image-instruction-label triplets as samples shown in Figure~\ref{fig:overview}(c) and more in the supplementary materials. We further built a subset including the attributes and phrases for chest X-ray images like "small base effusion, normal cardiac silhouette," which can be used as instruction for the report generation task. Additionally, the dataset underwent quality assurance by radiologists to ensure its accuracy and reliability. 
% Utilizing the multiple public datasets, we supplement it with segmentation results for pneumothorax, lung, and heart. 
% The dataset demonstrates high Dice Similarity Coefficient values ranging from 0.913 to 0.941, indicating a strong agreement between the GT and predicted masks. 

% \textbf{Ethical Statement}
% Our dataset consists of original images derived from publicly available datasets. It adheres to stringent ethical guidelines, as detailed in each dataset. 
% For instance, for the MIMIC-CXR dataset sourced from the United States, anonymization procedures were carried out in accordance with the Health Insurance Portability and Accountability Act of 1996 (HIPAA) Safe Harbor provisions. Access to this dataset necessitated the successful completion of a dedicated training course on human subjects research ethics. 
% The in-house dataset utilized for testing purposes was approved by the Ethics Committee of Fengcheng People's Hospital, Huanggang Hospital, and Longkou People's Hospital, and the committee waived the consent since the retrospective research will not change the examination process of the patients. All data were adequately anonymized.

\textbf{Instruction Design}
% In the clinical context of chest X-ray images, physicians typically identify potential diseases, locate relevant regions, and subsequently generate a comprehensive report based on observation. This process involves tasks such as disease classification, localization, and report generation. Historically, either multiple single-task models or a single multi-task model were employed to accomplish these goals, but these approaches lacked intrinsic correlations between tasks.
% By utilizing multiple instruction sets during the joint training approach, we not only enable the model to learn task-related features but also activate its potential capabilities to adapt to other tasks. 
To build and utilize multiple instruction sets (for each of four sub-tasks) during the joint training approach, we design a set of seed instructions with placeholders (later replaced with corresponding targets) to create diverse related task descriptions for coarse-grained task-level customization, such as samples illustrated in Figure \ref{fig:overview}(c). Following various instructions, our model can elegantly switch among different vision-centric tasks and accomplish them in a unified manner. More details about the organization of instructions for task-level customization, including disease classification, localization, segmentation, and report generation, are introduced in the supplementary material. 

% \textbf{Datasets for Training}
We employ a list of public datasets for training our proposed transformer model, e.g., MIMIC-CXR, VinDr-CXR, and ChestX-Det. 
We first sort out the aligned lesion categories of each dataset and the associated radiology report data and bounding box (BBox) data. We exclude the image datasets that are included in the test and validation datasets of downstream tasks to avoid data leakage. Each dataset is described in detail as follows:

\begin{itemize}
\item \textbf{MIMIC-CXR} \cite{johnson2019mimic} contains more than 377,110 radiograph images from over 227,835 radiographic studies. 
% Images in the MIMIC-CXR dataset are collected from multiple angled views. 
Each radiograph is paired with lesion classification and associated radiology report. We employ this dataset for multi-label classification and report generation tasks.
% For the report generation task, we extract the finding and impression sections of the report and remove redundant white space.
% following \cite{chen2020generating}. 
% Furthermore, we filter out irrelevant information, such as "compared with the previous report" and "discussed with doctors", the model can therefore focus on diagnosis-related information that can be obtained from images.

\item \textbf{Padchest} \cite{padchest} includes 160,868 images obtained from 67,625 patients, covering six different position views. 
% with related reports. The reports
It has 174 different radiographic findings and 19 differential diagnoses, totaling 193 classes. They are used for the classification task.

\item \textbf{CXR-AL14} \cite{CXR-AL14} is a large-scale dataset for chest X-ray image detection. It has more than 140,000 chest X-ray radiographs containing 253,844 bounding boxes in 14 chest abnormal object categories.
% (Atelectasis, Calcification, Consolidation, Effusion, Emphysema, Fibrosis, Fracture, Mass, Nodule, Pleural thickening, Pneumatosis, Pneumothorax, Postoperative metal, Venipuncture)
\item \textbf{VinDr-CXR} \cite{nguyen2022vindr} includes chest radiographs with annotations for the classification of 28 common chest diseases. The dataset contains 15,000 CXR scans in the training set. 
% Each scan is annotated by three radiologists. 
We select eight diseases from the dataset along with their corresponding BBox for the disease localization task.

\item \textbf{ChestX-Det} \cite{lian2021structure} consists of 3,578 images from NIH ChestXray14\cite{wang2017chestx} for 13 common disease.
% which are annotated by three radiologists  or abnormality categories. 
We select seven diseases from the dataset along with BBox for the disease localization task.

\item \textbf{CheXmask} \cite{cheXmask2023heart} contains 676,803 lung and heart segmentation masks of chest images from six publicly available databases: CANDID-PTX, ChestXray14, Chexpert, MIMIC-CXR, Padchest, and VinDr-CXR. We use 224,316 data for training and 10,000 data from ChestXray14 for downstream evaluation.

\item \textbf{SIIM} \cite{siim} comes from the SIIM-ACR Pneumothorax Segmentation competition and contains 12,090 images, among which approximately 3,000 cases are positive for pneumothorax disease with masks.

\item \textbf{In-house dataset} consists of 2,531 chest X-ray images, encompassing nine disease categories relevant to the disease localization task, along with BBox. All the images are captured from the front view.
\end{itemize}

\subsection{Model Architecture}
\begin{figure}[t]
	\centering
	%\fbox{\rule[-.5cm]{4cm}{4cm} \rule[-.5cm]{4cm}{0cm}}
    \includegraphics[width=0.95\linewidth]{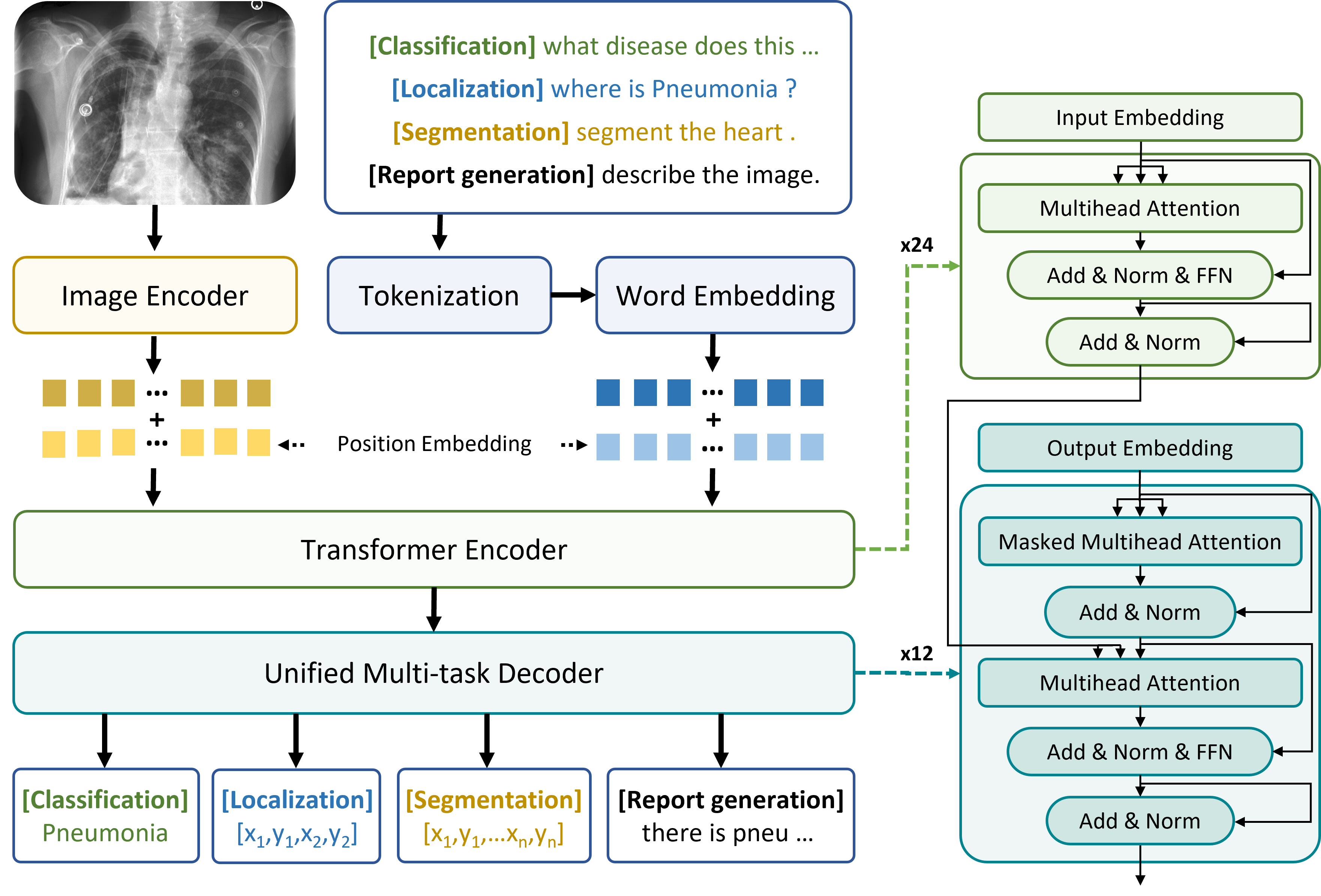}
	\caption{Overview of OmniFM-DR. Our model processes the chest X-ray image and instruction query utilizing a Resnet-based encoder and BPE/tokenizer, respectively. We then fuse the visual and textual information with a transformer encoder and generate the output text with a unified multi-task decoder. The instruction tuning includes four tasks, i.e., multi-disease classification, disease localization, segmentation, and report generation. 
}
	\label{fig:method}
\end{figure}

 % The attributes of the diseases  are extracted from the ground truth reports and used as customized instruction for report generation. Cardiotoracic Ratio (CTR) and Pneumothorax Compress Ratio (PCR) is obtained by post-processing the results of cardiac, lung and pneumothorax segmentation.
In this work, we propose a multimodal 
% multi-task 
model that leverages the 
% widely-used 
sequence-to-sequence learning paradigm for joint multi-task training. The specific tasks encompass disease classification, localization, segmentation, and report generation. For each task, we design specific task instructions to facilitate the model's differentiation between tasks. 
Inspired by the multi-modal models \cite{wang2022ofa,chen2022pix2seq,chen2022unified,lu2022unifiedio}, OmniFM-DR leverages an encoder-decoder architecture for perceiving pixel inputs and generating the target sequence and performs unified modeling and joint training on downstream visual and language tasks as shown in Figure \ref{fig:method}. Bounding boxes and class labels are converted into sequences of discrete tokens. This enables OmniFM-DR to robustly perform diverse language and vision tasks based on instructions, providing diverse and complex output results. 
% Benefiting from pre-trained language models and mutual guidance between tasks, OmniFM-DR can engage in continuous question-answering and provide visual explanations of answers during conversations, which is particularly crucial in safety-critical domains like medical diagnosis. 

% \subsection{Model Architecture}
\textbf{Image and Language Encoder} With an input chest X-ray image $ x_{i}  \in  \mathbb{R}^{H\times W} $, visual features are extracted by image encoder and further projected to the feature space:

\begin{equation}
    v_{i} = P_{img}(E_{img}(x_{i})) \in \mathbb{R}^{(h_{f} \times w_{f}) \times d}
\end{equation}

where $h_{f}$ and $w_{f}$ are the output size of visual features, and $d$ represents the feature dimension. $E_{img}$ can be any common visual backbones and we use ResNet152 in this study. Specifically, we take output features from the 4th residual block. Visual features are then projected to a pre-defined feature dimension by using $P_{img}$, which is composed of two linear layers.
% \textbf{Language and encoder } 
With any processed input instruction sequence $t_{i}$, text features are extraced by language encoder:

\begin{equation}
    l_{i} = E_{txt}(t_{i}) \in \mathbb{R}^{n_{t} \times d}
\end{equation}

where $n_{t}$ is the number of input tokens and $d$ represents the feature dimension. In our case, Bert \cite{devlin2018bert} is used as a language encoder.

\textbf{Multi-modality Module} This module follows an encoder-decoder architecture format. Given the input visual features $v_{i}$ and text features $l_{i}$, we first generate fused multi-modal representations by combining the image and text embeddings. These fused features serve as the keys and values in the cross-attention blocks in the decoder. By conditioning on the partial sequence $y_{i, <j}$ predicted so far, the decoder recursively makes predictions for the token at position $j$, effectively generating aligned descriptions across modalities. 

% The fused multi-modal embeddings capture relationships between the input modalities, guiding the decoder toward generating coherent descriptions grounded in both the image and text content.

\begin{equation}
    y_{i, j} = D_{mm}(E_{mm}(concat(v_{i}, l_{i})), y_{i, <j}) \in \mathbb{R}^{1 \times d}
\end{equation}

In our experiments, we leverage BART \cite{lewis2019bart} for multi-modal encoding and decoding. BART utilizes BERT \cite{devlin2018bert} as the encoder. The decoder is based on GPT \cite{radford2018improving} and generates output sequences in an autoregressive manner. 

\textbf{Model joint-training and Inference} We optimize the sequence-to-sequence model using cross-entropy loss as follows. 

\begin{equation}
    L = - \sum_{i=1}^{n}\sum_{j=1}^{|y|} \log P_{\theta}(y_{i, j} | y_{i, <j}, x_i, t_i)
\end{equation}
where $n$ is the batch size, $\theta$ represents the model parameters, $x_{i}$ represents the input image, $t_{i}$ stands for the input instruction, and $y_{i, j}$ denotes the output token at position $j$ for the $ith$ sample at each batch. To enhance the quality of generation during inference, we employ various decoding techniques, such as beam search.

\subsection{Downstream Finetuning and Evaluation}
The benchmark evaluates four tasks, using a total of nine datasets with over 140 thousand samples. Among them,
ChestX-ray14 and RSNA pneumonia datasets are utilized for evaluating the performance on the multi-label classification task, while 
MS-CXR, ChestX-ray14, and RSNA pneumonia datasets for the disease localization task.
Meanwhile, we assess the report generation on the MIMIC-CXR dataset.
Several standard metrics are introduced for various tasks. For example, F1 stands for “F1 score”, ACC stands for “Accuracy”, BLEU stands for “BiLingual Evaluation Understudy” \cite{papineni2002bleu}, ROUGE stands for “Recall-Oriented Understudy for Gisting Evaluation” \cite{lin2004rouge}. For BLEU and ROUGE, we all use 1-gram by default.

% \textcolor{black}{The statistical analysis was performed following the method in \cite{delong1988comparing,takahashi2022confidence}.}

We compared the proposed method with other methods across four tasks \cite{zhang2022contrastive,Huang2021GLoRIAAM,Boecking_2022,zhang2023knowledgeenhanced,wu2023medklip,Deng_2021_ICCV,Zhu_2022,VGTR2022vg,li2021referring,chen2023medical,anderson2018bottom,rennie2017self,chen2022crossmodal,liu-etal-2021-contrastive} and the detailed comparison results can be found in the supplementary materials. The evaluation datasets include

% CheXpert

% For the RSNA Pneumonia dataset, we randomly sampled 3,000 images for testing.
\begin{itemize}

% \item \textbf{ChestXray14} \cite{wang2017chestx} is an available dataset for diagnosing 14 common lung diseases and eight localization of key findings, with 984 radiograph images and hand-labeled BBox. We randomly split it into training/validation/test sets by 7:1:2 for the classification and localization task.

\item \textbf{ChestXray14} \cite{wang2017chestx} 
% is a publicly accessible dataset for classification and localization. It 
has 112,120 images with 14 common disease labels, with 984 images having eight localization of key findings with hand-labeled BBox. 
% We randomly split it into training/validation/test sets by 7:1:2 for the localization task.

% \item \textbf{CheXpert} \cite{irvin2019chexpert}contains 224,316 images collected from 65,240 patients. We extract 1\% of the dataset to conduct a finetuning experiment for multi-disease classification. We follow MRM to focus on 5 diseases: Atelectasis, Cardiomegaly, Consolidation, Edema, and Pleural Effusion. We sample training/test sets from the official training set, constituting 21,84/5,000 images of the whole dataset. For the CheXpert dataset, we randomly sample 5,000 images with the same 14 labels with the MIMIC-CXR dataset for the test. 

\item \textbf{MS-CXR} \cite{boecking2022making} consists of 1,153 samples with BBox and concise radiology reports for eight diseases (sourced from the testing set of MIMIC-CXR), which is utilized for the disease localization task.
% We randomly split it into training/validation/test sets by 7:1:2 based on the patients and evaluated the average performance of the model in all eight diseases. 
% MS-CXR dataset contains BBox information on eight diseases(i.e., Consolidation, lung opacity, Cardiomegaly, Pneumonia, Atelectasis, Edema, Effusion, and Pneumothorax).

\item \textbf{RSNA Pneumonia} \cite{shih2019augmenting} is a chest X-ray dataset for Pneumonia classification and localization. It comprises 26,683 images, where each radiograph is
categorized as either pneumonia or normal. 
% We randomly sample 3,000 data from the official
% training set to build the test set for direct inference and finetuning experiments for disease classification and localization tasks.

\item \textbf{JSRT} \cite{jsrt2000lung} contains 247 chest images, among which 154 cases have lung nodules.
% and 93 cases have no lung nodules, and was released by the Japanese Society of Radiological Technology. 
We selected healthy samples for the lung segmentation task.

\item \textbf{MS-PS} is a privately annotated dataset for pneumothorax segmentation, comprising 233 images from the MS-CXR dataset. 
% It was annotated by three expert physicians and is split into a 4:1 ratio for training and testing, respectively, for downstream pneumothorax segmentation evaluation.

% \item \textbf{Shenzhen} Chest X-ray set \cite{shenzhen2014lung} comprises a total of 662 frontal chest X-rays, obtained through collaboration with Shenzhen No.3 People's Hospital, Guangdong Medical College, Shenzhen, China. Among these X-rays, 326 cases are classified as normal, while 336 cases exhibit manifestations of tuberculosis (TB). We randomly split it into training/validation/test sets by 7:1:2 for the lung segmentation task.

\end{itemize}

\subsection{Clinical Evaluation of Generated Reports}
To examine the quality of generated reports from a clinical usefulness perspective, we conducted a comprehensive evaluation performed by three experienced radiologists.
A total of 160 cases were evaluated, including 120 cases from three medical institutes (listed in details below) and 40 cases from the MIMIC-CXR test set. To match the intended inputs of our model, we excluded cases that mentioned multiple imaging views or comparisons to prior test results in the generated reports.
Our study involved two distinct yet complementary human evaluations: (a) a parallel evaluation, where radiologists compared and ranked alternative reports based on their quality, and (b) an independent evaluation conducted to assess the quality of each individual report.

\textbf{Ethical Statement}
The private data used in this retrospective study was approved by the Ethics Committee of three institutes, i.e., Fengcheng People's Hospital, Huanggang Hospital of Traditional Chinese Medicine, and Longkou People's Hospital.
And the committees waived the consent since the retrospective research would not change the patients' examination process. All data were adequately anonymized, and the risk of disclosing patient privacy via imaging data was minimal.

\textbf{Parallel Evaluation} All 160 original and generated reports were randomly chosen from a pool of four centers and evaluated by three radiologists. The radiologists were unaware of the source of the reports and reviewed them in a randomized order.
The quality of the reports will be scored subjectively on a 1-5 scale, with 1 being the worst and 5 the best. The detailed guidelines are provided as follows:

% \vspace{-3pt}
% \begin{quote}
% \setlength{\leftskip}{-3em}
\begin{itemize}
% \item Reports with fewer errors will be selected;
% \item Reports with fewer errors (e.g. missing or incorrect description) will be prioritized;

\item Report without diagnosis errors and with comprehensive description should be scored 5;

\item Report with significant clinical diagnosis errors should be scored 1;

\item Report can be considered correct if the content is reasonable based on the given image. For instance, an accurate diagnosis of pleural effusion may not be obtained based on a frontal view image;

\item Report can be considered correct if there are descriptions of related diseases. For instance, lung collapse can be indicative of atelectasis;

\item Report with the better description should be scored higher, if two reports are error-free or exhibit similar errors;

\item Repetitive descriptions can be overlooked;
\end{itemize}

\textbf{Independent Evaluation} Radiologists were provided with one chest X-ray image paired with the disease findings and tasked with assessing the generated reports and original reports. During the evaluation, the radiologists were unaware of the source of the reports. They aimed to determine whether there are discrepancies or errors, any missing elements, or inaccurate descriptions (e.g., location and severity) in the reports and evaluate their clinical significance referring to the methodology \cite{yu2022evaluating, xu2023elixr}.
Six types of diseases are evaluated, i.e., Pneumothorax, Pleural Effusion, Edema, Consolidation or Pneumonia (grouped together), Atelectasis, and \textcolor{black}{N}ormal.
% following\cite{xu2023elixr}. 
Radiologists were required to assess whether every type of error exists for each specific disease when evaluating reports.
The considered errors are agreed by the radiologists and listed as follows:

% \vspace{-3pt}
% \begin{quote}
% \setlength{\leftskip}{-3em}
\begin{itemize}
\item False positives. Incorrect disease detection;
% \item False positives: Incorrect disease detection. For instance, pneumonia is mentioned in the report when it is not present, or pneumonia is identified as another disease;

\item False negatives. Missed disease detection;
% \item False negatives: missed disease detection. For instance, pneumonia is present but not mentioned or described in the report;
    
\item Inaccurate location. For instance, left lung pneumonia is described as right lung pneumonia;
    
\item Inaccurate severity. For instance, a minor pleural effusion is described as a major pleural effusion;
    
\item Non-existent references. For instance, "compared with previous" and "based on front-lateral image";

\end{itemize}

\textcolor{black}{\subsection{Statistical Analysis}
We conducted five training iterations of the model using different random seeds and recorded its performance on all tasks each time \cite{zhou2023foundation}. We calculated the mean and standard deviation of the model's performance and obtained a 95\% confidence interval through 
% \( \text{E} \pm 1.96\sigma/\sqrt{5} \)
\( mean \pm 1.96 \times standard \: deviation /\sqrt{5} \).
For tasks with multiple categories, such as multi-class classification or multi-class localization, we first calculated the performance for each category and then averaged them to obtain an overall performance measure. In the clinical experiment section, the side-by-side comparison analysis was conducted using a Chi-Square test to verify that there was no significant difference between the generated reports by the model and the reports by doctors. The independent evaluation analysis was performed by generating 1000 bootstrap samples and reporting the 2.5th and 97.5th percentiles as the 95\% confidence interval.}

\section*{Code and Data Availability}
Code for training and evaluation is available at \url{https://github.com/MedHK23/OmniFM-DR}.
The new dataset released in this study can be found at \url{https://huggingface.co/datasets/MedHK23/OmniFM-DR}.
The MultiMedBench is all open source, and the respective download link is described in Git Hub.

% \section*{Acknowledgements}
% This work was supported by the program.

\section*{Author contributions}
All authors have contributed fully to the concept and design of the study. LX and ZN collected the clinical data, performed the experiments, and analyzed the experiment results. XL performed the comparative experiments with other methods. LX and XW drafted the manuscript. XW, SZ, and HL supervised the projects and gave final approval of the manuscript. All authors have carefully read and approved the final manuscript.

\section*{Competing interests}
The authors declare no competing interests.

% Correspondence and requests for materials should be addressed to Lijian Xu or Xiaosong Wang.

\bibliographystyle{unsrt}
\bibliography{references}  %%% Uncomment this line and comment out the ``thebibliography'' section below to use the external .bib file (using bibtex) 


% \end{thebibliography}

\clearpage

% 注释supplement

\section*{Supplementary Material}
\renewcommand{\tablename}{Supplementary Table}
\setcounter{table}{0}

\subsubsection*{Instruction Design}
In the clinical context of chest X-ray images, physicians typically identify potential diseases, locate relevant regions, and subsequently generate a comprehensive report based on observation. This process involves tasks such as disease classification, localization, and report generation. Historically, either multiple single-task models or a single multi-task model were employed to accomplish these goals, but these approaches lacked intrinsic correlations between tasks.
By utilizing multiple instruction sets during the joint training approach, we not only enable the model to learn task-related features but also activate its potential capabilities to adapt to other tasks. 
As described in Figure \ref{fig:overview}(c), we develop a series of instructions containing placeholders, allowing us to generate a wide range of task descriptions for high-level task-specific modification. By adhering to these diverse instructions, our model is capable of smoothly transitioning between different vision-based tasks and executing them in a harmonized fashion. Here, we introduce the organization of instructions for task-level customization, including disease classification, localization, segmentation, and report generation as follows. 
% For example,

\textbf{Disease Classification Dataset} includes entity information across 193 categories and 0.54M images.
For the entity classification task, the instruction is "What disease does this image have?". The answer includes all possible diseases present in the data, such as "pneumonia" and "atelectasis.", "Is Pneumonia in this image?". The response can be either "yes" or "no". 
% In order to develop an attribute subset, 
We further extracted the textual phrases from the disease attributes (e.g., small left pneumothorax, normal cardiac silhouette) described in the original report of MIMIC-CXR and developed a subset that matches 135,751 images with phrases. 
The subset comprises position descriptions (e.g., left, right, base, mid) and severity descriptions (e.g., mild, moderate, severe) for ten common diseases, i.e., Cardiomegaly, Pneumonia, Effusion, Atelectasis, Edema, Consolidation, Pneumothorax, Opacity, Fracture, and Supported Devices.
% The inclusion of  are highly beneficial for improving the quality of reports. 
% During the inference stage, these phrases are obtained from the classification tasks and disease localization tasks and subsequently utilized as customized instructions for report generation.
For the severity classification task, the instruction is "What is the level of cardiomegaly?". The response can be "moderate" or "severe". The instruction for the location classification task is like "Where is pneumothorax?". The response can be "on the left apical side". 
 
\textbf{Disease Localization Dataset} 
 incorporates CXR-AL14, VinDR-CXR, ChestX-Det, and In-house datasets, consisting of 0.19M images and corresponding BBox for 12 diseases.
 The instruction given for the disease localization task is "Give the accurate bounding box of \{\}.". Here, the placeholder \{\} represents the category of the specific disease, such as "pneumonia, in the lower left lung". The response is a distinct bounding box area defined by coordinates [x\textsubscript{1}, y\textsubscript{1}, x\textsubscript{2}, y\textsubscript{2}], representing the top-left and bottom-right points.
% \textbf{The pneumothorax \&  cardiac \& lung segmentation subset} includes contour points (polygon vertexes, recomputed from the region mask) for cases of pneumothorax. We get preliminary segmentation results of 233 pneumothorax cases by utilizing OmniFM-DR (trained based on ????) from MS-CXR, which has pneumothorax BBox for reference. Subsequently, we submit the segmentation results to the annotation platform and seek rapid judgments from three qualified doctors (FOR WHAT PURPOSE????). Additionally, we supplement the SIIM positive data by converting it into the appropriate format and thus get a total of 2,717 pneumothorax samples in the dataset.

 \textbf{Segmentation Dataset} includes CheXmask and SIIM datasets for the segmentation task, comprising 0.23M images. 
 % \textbf{The cardiac \& lung segmentation subset} includes contour points of cardiac and lung segmentation. 
 % We use the CheXmask dataset for pretraining and employ OmniFM-DR to inquire about the precise locations of the heart and left/right lungs in all the image data from MIMIC-CXR. 
 % Subsequently, 
 We calculate the Cardiothoracic Ratio (CTR) for each image and compare it with the corresponding relationship described in the reports (e.g., CTR < 0.51: normal cardiac silhouette; 0.51 < CTR < 0.55: mild cardiomegaly; 0.55 < CTR < 0.6: moderate cardiomegaly; CTR > 0.6: severe cardiomegaly). This comparison allows us to filter the data accordingly. 
 The SIIM dataset is collected for pneumothorax segmentation. We further supplement the disease phrase subset and segmentation subsets as follows. The pneumothorax subset includes contour points (polygon vertexes, recomputed from the region mask) for 233 cases of pneumothorax.
 % Regarding the segmentation task, 
 The respective instruction is "Please segment the \{\} from the given image." For instance, "Please segment the heart from the image." The response is a polygon area defined by a set of 30 points (coordinates).

\textbf{Report Instruction Dataset} 
includes the original MIMIC-CXR dataset of 0.24M front images, and paired radiology reports.
The instruction provided for the report generation task is "describe the image". This task specifically involves generating comprehensive reports based on chest X-ray images. 
Such brief instruction generates reports that lack accurate descriptions. We thus incorporate disease attributions in the instruction to improve the quality of the reports. During the training stage, we extract disease entities from ground truth reports and relevant severity and position attributes of the diseases within the corresponding sentences. These attributes are then combined with the original instruction for training. During the inference stage, we construct instructions for report generation using the results of the classification, segmentation, and disease localization tasks. First, we obtain the disease category from the classification task. Then, we use disease localization to determine the location and size of the lesion and compare it with the lung mask to determine the precise positional information.

\subsubsection*{\textcolor{black}{Training and} Experiment Details}
In this section, we introduce the \textcolor{black}{training procedure,} detailed setting of the direct inference, and fine-tune across all four tasks. \textcolor{black}{Throughout the training process, our methodology guarantees the inclusion of data from all tasks in each batch. Each training sample within the batch is composed of task-specific instructions and accompanying images. These instructions play a crucial role in facilitating the model's ability to discern between various tasks and generate corresponding outputs. Additionally, the computation of the loss function takes place concurrently within each batch.}
% Several standard metrics are introduced for various tasks.For example, F1 stands for “F1 score”, ACC stands for “Accuracy”, BLEU stands for “BiLingual Evaluation Understudy” \cite{papineni2002bleu}, ROUGE stands for “Recall-Oriented Understudy for Gisting Evaluation” \cite{lin2004rouge}. For BLEU and ROUGE, we all use 1-gram by default. 
 Based on empirical findings, we set the proportional distribution of training data across each batch for classification, disease localization, report generation, and segmentation tasks to be 0.15/0.2/0.5/0.15. 
 % For the disease localization task, we select the data with the same disease labels as the MIMIC-CXR dataset on each dataset to join the training. 14 lesions in the MIMIC-CXR dataset are regarded as the standard for the disease classification task. Meanwhile, the accurate delineation of three class target regions (e.g., pneumothorax, lung, and heart) is completed by the segmentation task. 
 All the images are resized to a uniform size of 512x512 and subsequently adjusted by contrast and brightness. We selected the huge version of the OFA model \cite{wang2022ofa} as the pre-training model. We set a learning rate of $10^{-5}$, warm-up learning rate of  $10^{-7}$, and dropout rate of 0.1, and train on eight V100 with batch size 256 for 30 epochs. \textcolor{black}{During inference, we set a beam width of 6.}
 % We fine-tune all models using a learning rate of  $10^{-4}$  for all datasets with a batch size of 64. 

• \textbf{Classification and Segmentation} 
%         &\#MedKLIP\cite{wu2023medklip}             & 76.7 & 28.8  & 86.9 & 63.4 &  & -\\
%         % & KAD\cite{zhang2023knowledgeenhanced}   & 79.6 & 31.9 & - & - & - & -\\
We have selected ConVIRT\cite{zhang2022contrastive}, GLoRIA\cite{Huang2021GLoRIAAM}, and BioVil\cite{Boecking_2022} as the baseline models for both disease classification and segmentation task. For classification, we strictly follow the official train-test split for ChestXray14 and MedKLIP\cite{wu2023medklip} test setting for RSNA, respectively. For segmentation, we randomly split the dataset into training/validation/test sets by 7:1:2. In all three models, ResNet-50 and BERT are chosen as the visual and text encoders, respectively. To perform direct inference of classification, we adopt the methods proposed in GLoRIA and BioVil, which transform the image classification task into a text-image matching task. Specifically, the test image is fed into the image encoder to generate image features. The test disease labels are subsequently formulated as text prompts and fed into the text encoder to generate text features. We then calculate the similarity between the image and text features. The prediction scores are set with normalized similarities. For ACC and F1, we utilize the validation dataset to determine the best score threshold for each class. Furthermore, we adhere to the official training strategies and train each model for 50 epochs during the fine-tuning process. Supplementary Table~\ref{tab:cls} shows the classification tasks achieve satisfactory results across all diseases. \textcolor{black}{For disease attribute classification, we considered combining synonymous location descriptions. When calculating the metrics, we treated "lower lobe/base/basal/basilar" as "lower" and "apical/upper" as "upper." And we enhanced disease classification by considering vocabulary related to disease uncertainty in the reports\cite{zhang2023expert}. For instance, we treated low-probability terms like 'not exclude' and 'cannot accurately assess' as negative indicators, while high-probability terms like 'likely' and 'probable' were considered positive indicators. Additionally, we extended the keyword search for describing disease severity. For example, terms like 'severe', 'moderate to severe', and 'moderate to large' all indicated 'severe' conditions.}
% The results are comparable with KAD\cite{zhang2023knowledgeenhanced} and MedKLIP\cite{wu2023medklip} claimed SOTA.
 % KAD claimed SOTA in several metrics, which can not be reproduced due to the lack of the code.
% Supplementary Table~\ref{tab:cls} show that the classification results are slightly inferior to SOTA models since the model proposed is based on generative artificial intelligence. When the corresponding classification heads are added, the performance of classification and segmentation tasks also achieves satisfactory results on both direct inference and Few-shot across several datasets (shown as ours*). Since most models have been focusing on natural images and there do exist feature gaps between natural and medical images. We have fine-tuned corresponding models using 100 and full samples. 

• \textbf{Disease Localization} 
We employ TransVG, SeqTR, and VGTR as the baseline models for comparison with our OmniFM-DR.
Due to the absence of an official training/validation/test ratio for the disease localization datasets, we randomly split it into training/validation/test sets by 7:1:2 for the localization task.
Take note that the MS-CXR and ChestXray14 dataset doesn't come with an official division ratio for training/validation/test in the disease localization task, and the RSNA Pneumonia dataset is devoid of ground truth bounding boxes for their test sets. As a response, we randomly partition the official training sets of these datasets into training/validation/test sets adhering to a 7:1:2 ratio for the upcoming fine-tuning experiments.
For TransVG\cite{Deng_2021_ICCV}, Resnet-50 is selected as the backbone. The BERT and ViT encoding length are 12 and 6 separately, while the maximum query length is set to 20, following the authors' recommendation. TransVG\cite{Deng_2021_ICCV} has been trained on five datasets and all models are validated, while the most competitive on RefCOCOg is reported in Supplementary Table~\ref{tab:VG}. 
For SeqTR\cite{Zhu_2022}, we follow the default settings of RefCOCOg, and DarkNet53 is selected as the detection backbone. The corresponding pre-calculated word embeddings are used to accommodate the pre-trained models. The authors have released three models on different datasets and training settings. We validate each model and the most competitive on RefCOCOg is reported in Supplementary Table~\ref{tab:VG}. For VGTR\cite{VGTR2022vg}, we followed the default settings as RefCOCOg and selected ResNet-50 and Bi-LSTM as the vision backbone and text encoder, respectively. For the evaluation of disease localization, the IoU threshold of TransVG, SeqTR, VGTR, and OmniFM-DR is set as 0.5 consistently. 
% The results of RefTR\cite{li2021referring} and MedRPG\cite{chen2023medical} can not be reproduced due to the absence of code.
% Supplementary Table~\ref{tab:VG} shows that OmniFM-DR consistently achieved SOTA performance in the tests of direct inference, 100-shot and full-data finetuning.

• \textbf{Report Generation}
We utilize Up-down\cite{anderson2018bottom}, Att2in\cite{rennie2017self}, and R2GenCMN\cite{chen2022crossmodal} as the baseline models for report generation. The official MIMIC-CXR test dataset is used for evaluation. Both Up-down and Att2in employ LSTM as the text encoder. Following their official implementation, Faster R-CNN and ResNet-101 are chosen as the image encoders for Up-down and Att2in, respectively. For R2GenCMN method, ResNet-101 serves as the image encoder, while a transformer-based module is utilized as the language model. 
% The number of memory vectors is set to 2048, with a dimension of 512. In terms of data preprocessing, we extract the finding and impression sections of the report and remove redundant white space, as outlined in \cite{chen2020generating}. Moreover, we filter out irrelevant information, such as phrases like "compared with the previous report" and "discussed with doctors", focusing on diagnosis-related information that can be obtained from the images. 
We retrain the three models following their training procedures with our preprocessed dataset and evaluate them on the official test dataset. As for OmniFM-DR, leveraging its multi-task capability, we find it beneficial to incorporate disease attributes as prompts during both the training and inference stages. During training, we include extracted phrases from radiologist reports as additional prompts. During inference, we utilize phrases predicted by our model as supplementary prompts. Supplementary Table~\ref{tab:ablation} shows that our report generation task achieved SOTA on clinical efficacy metrics and comparable results on natural language processing metrics. 
Supplementary Figure \ref{fig:examples} provides more examples of multi-task results generated by our model. It can be found that the proposed model is capable of identifying Pneumothorax(a), Pneumonia (b), Edema (c), and Atelectasis(d) with a disease localization box, classification, and generated report. 
Take Supplementary Figure \ref{fig:examples}(b) for example, 
the generated report demonstrates the accurate pneumonia features and position described as "increased opacification of the bilateral bases, right greater than left", which are well consistent with the blue highlighted text in the golden standard report. The disease localization and classification results also agree with the gold standard. 
Furthermore, the generated report shows a stable cardiomediastinal contour which could be verified by the cardiothoracic ratio of 0.49 calculated by the segmentation task. 
Through the validation of multi-tasks, the explainability of the generated reports could be greatly enhanced.

\begin{table}[]
\caption{Comparison with other state-of-the-art methods of \textbf{Disease classification} task with direct inference  setting on ChestXray14 dataset. The mean metrics (e.g., AUC and F1) refer to the macro average on the 14 diseases. }
    \centering
    % \resizebox{\linewidth}{!}{
    % \scalebox{0.8}{%
    \begin{tabularx}{\textwidth}{Xc*{15}{X}}
    % \begin{tabular}{ccccccccccccccccc}
\hline
  % Dataset& 
 \rotatebox{90}{Metric} & Model & \rotatebox{90}{Mean}& \rotatebox{90}{Atelectasis} & \rotatebox{90}{Cardiomegaly} & \rotatebox{90}{Effusion} & \rotatebox{90}{Infiltration} & \rotatebox{90}{Mass} & \rotatebox{90}{Nodule}  &\rotatebox{90}{Pneumonia} &\rotatebox{90}{Pneumothorax} &\rotatebox{90}{Consolidation} &\rotatebox{90}{Edema} & \rotatebox{90}{Emphysema} & \rotatebox{90}{Fibrosis} & \rotatebox{90}{Pleural Thicken} & \rotatebox{90}{Hernia} \\
\midrule
% ChestXray14&
% 0.560 0.459 0.433 0.646 0.654 0.601 0.580 0.640 0.533 0.646 0.692 0.431 0.482 0.545 0.494
AUC &  ConVIRT\cite{zhang2022contrastive}     & 56.0 & 45.1 & 44.3 & 63.2 & 65.1 & 61.6 & 57.2 & 63.6 & 54.1 & 63.7 & 70.2 & 41.9 & 47.4 & 56.0 & 51.2\\
                 &   GLoRIA \cite{Huang2021GLoRIAAM} & 60.9 & 65.9 & 70.2 & 74.9 & 65.2 & 60.9 & 52.1 & 59.4 & 56.6 & 69.3 & 74.4 & 50.1 & 46.2 & 60.7 & 46.0 \\
                &  BioViL\cite{Boecking_2022}       & 65.9 & 52.1 & 68.4 & 74.4 & 61.2 & 65.4 & 63.4 & 67.2 & 68.3 & 64.1 & 78.1 & 64.4 & 62.0 & 64.8 & 69.0 \\
                & MedKLIP\cite{wu2023medklip}& 72.4 & 65.3 & \textbf{84.9} & \textbf{82.3} & \textbf{68.2} & \textbf{74.6} & 64.2 & \textbf{69.9} & 79.4 & 70.2 & 79.3 & \textbf{79.6} & 58.6 & 51.2 & 85.9 \\
                 &Ours & \textbf{73.7} & \textbf{72.8} & 76.5 & 78.1 & 62.8 & 73.9 & \textbf{64.2} & 69.7 & \textbf{82.6} & \textbf{71.4} & \textbf{81.5} & 73.1 & \textbf{63.1} & \textbf{73.3} & \textbf{88.6}\\
 F1&  ConVIRT\cite{zhang2022contrastive}            & 13.4 & 0.6 & 0.4 & 36.4 & 43.4 & 15.5 & 14.0 & 5.8 & 20.3 & 17.5 & 12.3 & 8.1 & 3.2 & 9.9 & 0.5 \\
                 &   GLoRIA \cite{Huang2021GLoRIAAM} & 17.5 & 28.4 & 15.2 & 46.1 & 46.6 & 13.2 & 12.6 & 5.9 & 21.4 & 21.2 & 13.9 & 8.1 & 0.3 & 11.2 & 0.8 \\
                 &  BioViL\cite{Boecking_2022}       & 19.2 & 23.5 & 20.9 & 43.8 & 41.4 & 17.8 & 16.4 & 6.7 & 27.4 & 17.7 & 18.7 & 12.3 & 5.6 & 11.9 & 4.5 \\
                & MedKLIP\cite{wu2023medklip}& 24.3 & 28.3 & \textbf{31.4} & \textbf{50.8} & \textbf{48.9} & 24.1 & 17.5 & 7.2 & 42.1 & 21.4 & 19.1 & \textbf{24.1} & 7.3 & 1.4 & 16.3 \\
                 &      Ours                              & \textbf{26.4} & \textbf{35.7} & 22.1 & 47.2 & 41.3 & \textbf{31.4} & \textbf{19.3} & \textbf{8.2} & \textbf{44.9} & \textbf{22.7} & \textbf{20.6} & 16.5 & \textbf{8.3} & \textbf{18.4} & \textbf{32.9}\\

		\bottomrule
    \hline
\end{tabularx}
% \end{tabular}%
% }
    \label{tab:cls}
\end{table}

   \begin{table}[]
\caption{Comparison with other state-of-the-art methods of \textbf{Disease localization} task with 20-shot setting on MS-CXR and ChestXray14 dataset. The metrics (i.e. ACC and mIoU)  refer to the macro average on the eight diseases.}
    \centering
    % \resizebox{\linewidth}{!}{
    \begin{tabularx}{\textwidth}{*{3}{c}*{12}{X}}
    % \begin{tabular}{ccccccccccccccc}
\hline
  %Lung Opacity&&  Consolidation
  % Mass, Infiltration, Cardiomegaly, Pneumonia, Nodule, Atelectasis, Effusion, Pneumothorax
  Dataset&Metric & Model & Mean & \rotatebox{90}{Atelectasis} & \rotatebox{90}{Cardiomegaly} & \rotatebox{90}{Effusion}  &\rotatebox{90}{Pneumonia} &\rotatebox{90}{Pneumothorax} &\rotatebox{90}{Consolidation} &\rotatebox{90}{Edema} &\rotatebox{90}{Opacity} &\rotatebox{90}{Infiltrate} &\rotatebox{90}{Mass} &\rotatebox{90}{Nodule}\\
  % &\rotatebox{90}{Infiltration} &\rotatebox{90}{Mass}&\rotatebox{90}{Nodule}\\
  % &\rotatebox{90}{Emphysema}&\rotatebox{90}{Pleural Thicken}\\
\midrule
 MS-CXR&ACC & VGTR\cite{VGTR2022vg}   &27.1	&30.0	&88.2	&9.3	&32.4	&9.8	&22.1	&11.7	&13.0	&-	&-	&-
\\
                 % &   REFTR\cite{li2021referring}     &&  & &  &&-\\
                 &&   TransVG\cite{Deng_2021_ICCV} &27.7	&19.0	&79.3	&23.9	&32.5	&9.3	&26.1	&15.0	&16.7	&-	&-	&-\\
                 && SeqTR\cite{Zhu_2022}&45.8	&30.2	&86.2	&13.9	&42.8	&\textbf{56.3}	&\textbf{46.4}	&37.0	&53.6	&-	&-	&-\\
                 % & MedRPG\cite{chen2023medical}  &&  &  &  &&-\\
                 &&	\textbf{Ours} &\textbf{55.4}	&\textbf{43.0}	&\textbf{94.1}	&\textbf{46.7}	&\textbf{59.9}	&45.0	&45.6	&\textbf{52.5}	&\textbf{56.6}	&-	&-	&-\\
       &mIoU & VGTR\cite{VGTR2022vg}&25.7	&30.3	&60.8	&16.5	&29.8	&12.1	&23.7	&15.8	&16.6	&-	&-	&-\\
                 % &   VGTR\cite{li2021referring}     &&  & &  &&-\\
                 &&  TransVG\cite{Deng_2021_ICCV}&27.6	&23.4	&60.2	&23.7	&28.8	&15.7	&25.4	&19.8	&24.1	&-	&-	&-\\
                 && SeqTR\cite{Zhu_2022} &44.8	&31.2	&\textbf{73.5}	&16.7	&45.4	&\textbf{55.9}	&47.3	&44.8	&43.4	&-	&-	&-\\
                 % & MedRPG\cite{chen2023medical}  &&  &  &  &&-\\
                 &&	\textbf{Ours}&\textbf{50.8}	&\textbf{47.0}	&69.8	&\textbf{43.9}	&\textbf{54.0}	&46.3	&\textbf{47.8}	&\textbf{50.3}	&\textbf{47.7}	&-	&-	&-\\
		% \bottomrule
 ChestXray14&ACC & VGTR\cite{VGTR2022vg} &28.9	&17.9	&86.4	&22.3	&41.3	&5.5	&-	&-	&-	&34.9	&12.9	&9.7\\
                 % &   VGTR\cite{li2021referring}     &&  & &  &&-\\
                 &&  TransVG\cite{Deng_2021_ICCV} &29.3	&9.6	&96.2	&10.2	&48.4	&15.8	&-	&-	&-	&36.0	&12.4	&5.9\\
                 && SeqTR\cite{Zhu_2022}&50.7	&47.2	&94.0	&\textbf{47.9}	&45.4	&\textbf{60.4}	&-	&-	&-	&58.2	&42.3	&10.4\\
                 % & MedRPG\cite{chen2023medical}  &&  &  &  &&-\\
                 &&	\textbf{Ours}&\textbf{60.9}	&\textbf{51.8}	&\textbf{99.4}	&46.6	&\textbf{62.1}	&48.3	&-	&-	&-	&\textbf{63.3}	&\textbf{64.3}	&\textbf{51.3}\\
       &mIoU & VGTR\cite{VGTR2022vg}&29.2	&22.1	&57.8	&29.5	&44.7	&12.3	&-	&-	&-	&37.7	&20.4	&9.5\\
                 % &   VGTR\cite{li2021referring}     &&  & &  &&-\\
                 && TransVG\cite{Deng_2021_ICCV} &29.9	&20.7	&69.9	&23.2	&42.5	&21.5	&-	&-	&-	&34.4	&17.9	&8.9\\
                 &&SeqTR\cite{Zhu_2022}&46.9	&44.1	&\textbf{80.4}	&39.1	&46.2	&\textbf{57.5}	&-	&-	&-	&53.3	&35.6	&19.0\\
                 % & MedRPG\cite{chen2023medical}  &&  &  &  &&-\\
                 &&	\textbf{Ours}&\textbf{51.3}	&\textbf{48.0}	&71.5	&\textbf{44.0}	&\textbf{54.8}	&46.7	&-	&-	&-	&\textbf{55.6}	&\textbf{52.0}	&\textbf{38.0}\\
		\bottomrule
\hline
% \end{tabular}}
\end{tabularx}
    \label{tab:VG}
\end{table}

\begin{table}[]
    % \caption{Ablation study of OmniFM-DR by removing or replacing individual modules. OmniFM-DR w/o classification uses only image features as key or value when training report generation model. OmniFM-DR w/o CTR only trains the models when training report generation model.}
    \caption{Ablation experiment of multi-task and prompt capability. The quality of the generated report is evaluated by report, entity, and attribute levels, with the overall performance assessed by metrics (i.e., BL-4, METEOR, and Rouge-L), and the accuracy of the disease category evaluated by the CE metric (i.e., Precision, Recall, F1). The attribute metric focuses on the performance of disease severity and location described in the report.}
        \centering
        \begin{tabular}{cccccccccc}
    \hline
    % \multirow{2}{*}{} 
     \multicolumn{2}{c}{}
     & \multicolumn{3}{c}{Report}
     & \multicolumn{3}{c}{Entity}
     & \multicolumn{2}{c}{Attribute}
     \\
     % &&  & PHRASE &  &   ENTITY& &  & REPORT &  \\
     &&  BL-4 & METEOR & Rouge-L  & Precision &Recall & F1  &ACC\_S & ACC\_L \\
    % \midrule
    % Method&Up-down\cite{anderson2018bottom}   &9.1	&12.8	&26.3	&32.2	&23.4	&23.9 &\\
    % &Att2in\cite{rennie2017self} 	   &9.7	&13.6	&27.5	&32.5	&23.6	&25.7 &\\
    % Method&R2GenCMN\cite{chen2022crossmodal}  &10.4&\textbf{14.7}	&\textbf{28.1}	&33.3	&28.4	&28.6 &\\
    % &\#Constrastive\cite{liu-etal-2021-contrastive} &10.9&15.1&28.3	&35.2	&29.8	&30.3 & \\
    % &\#KiUT\cite{huang2023kiut}         &11.3&16.0	&28.5	&37.1	&31.8	&32.1 &\\
   % &\#Med-PaLM M\cite{tu2023towards}    &11.5 & - & 27.5	&-	&-	&39.8 & \\
 \midrule
     % Baseline&Ours  & 10.97 & 14.02  & 26.48  & 43.22 & 31.31 & 33.29 & 18.34 & 8.17\\
     Baseline&Ours  & 11.12 & 14.08  & 26.53  & 42.77 & 31.49 & 33.08 & 18.82 & 8.31\\
    \midrule
    Task &- LOC    & 10.90 & 14.02 & 26.51 & 44.89 & 31.01 & 32.96 & - & - \\
         &- CLS    & 10.84 & 13.91 & 26.39 & 46.35 & 30.45 & 32.76 & - & - \\
    \midrule
   % Prompt &+ Entity                & 10.08 & 13.48  & 24.04 & 34.46 & 36.85 & 32.82 & 13.03 & 13.32 \\
   % Prompt &+ Phrase     & 10.22 & 13.65  & 24.42 & 35.71 & 38.11 & 35.08 & 22.19 & 12.87 \\
   %        &+ Phrase-GT             & 11.42 & 14.33  & 26.99 & 71.47 & 44.82 & 49.55 & 30.18 & 23.57  \\
    Prompt &+ Phrase     & 10.43 & 13.82  & 24.69 & 37.12 & 36.17 & 36.32 & 24.05 & 16.78 \\
          &+ Phrase-GT             & 11.35 & 14.21  & 27.42 & 70.69 & 46.04 & 50.41 & 31.93 & 24.64  \\
          
%           &w Entity\&Attribute-GT  & 11.74 & 14.91  & 27.92 & 73.10 & 45.15 & 50.90 & - & -\\
%          && BL-4 & METEOR & Rouge-L & Precision &Recall & F1 & ACC1 & ACC2 \\
%     \midrule
%     Tasks &OmniFM-DR & 10.97 & 14.02  & 26.48 & 42.52  & 31.36 & 33.29 & - & - \\
%     &w/o VG task     & 10.85 & 14.06  & 26.49 & 44.96  & 31.06 & 33.01 & - & - \\
%     &w/o CLS task    & 10.81 & 13.93  & 26.43 & 46.42  & 30.37 & 32.73 & - & -\\
%     \midrule
%    Prompt &w Entity     & 10.08 & 13.65  & 24.04 & 34.46 & 36.85 & 32.82 & - & -\\
% &w Entity\&Attribute     & 10.22 & 13.48  & 24.42 & 35.71 & 38.11 & 35.08 & - & -\\
%           &w Entity-GT  & 11.42 & 14.33  & 26.99 & 71.47 & 44.82 & 49.55 & - & -\\
%           &w Entity\&Attribute-GT  & 11.74 & 14.91  & 27.92 & 73.10 & 45.15 & 50.90 & - & -\\
          
    % \midrule
    % multi-disease&w CTR prompt & 11.02 & 14.03  & 26.53 & 42.73  & 32.04 & 33.43 & - \\
    % multi-disease&w PCR prompt & 10.57 & 13.86  & 26.48 & 41.92 & 31.36 & 33.29 & - \\
    % Cardiomegaly &wo prompt* & - & -  & - & 69.65  & 57.03 & 62.71 & 45.55 \\
    % Cardiomegaly &w  prompt* & - & - & - & 70.53  & 62.82 & 65.61 & 63.07 \\
    % Pneumothorax &wo prompt* & - & - & - &  75.38 & 66.12 & 70.45 & \\
    % Pneumothorax &w  prompt* & - & - & - &  74.92 & 68.62 & 71.37 & \\
    \bottomrule
    \hline
    \end{tabular}
        \label{tab:ablation}
    \end{table}
    
\begin{table}
	\caption{Diagnostic accuracy comparison with various \textbf{report generation} methods on MIMIC-CXR.
 % \# denotes the published methods that cannot be reproduced due to the lack of code. The corresponding metrics are cited from their paper.
 }
	\centering
    \begin{tabular}{cccccccccc}
		\toprule
		\cmidrule(r){1-9}
		Dataset &Model &BL-1 &BL-4  &METEOR & Rouge-L & Precision & Recall & F1\\
		\midrule
  MIMIC-CXR &Up-down\cite{anderson2018bottom}   &31.3 &9.0&12.7&26.1&32.4&24.1&24.0\\
  &Att2in\cite{rennie2017self}&33.4 &9.8	&13.8	&27.9	&31.9	&24.3	&25.5\\
  &R2GenCMN\cite{chen2022crossmodal}&35.1 &10.2&14.5&28.3&34.7&27.6&28.9\\
  &Constrastive\cite{liu-etal-2021-contrastive} &35.4&10.9&15.3&\textbf{28.4}	&36.3	&30.1	&30.6 \\
  % &\#KiUT\cite{huang2023kiut}                   &39.3 &11.3	&16.0	&28.5	&37.1	&31.8	&32.1 &\\
  % &\#Med-PaLM M\cite{tu2023towards}             &32.1 &11.5 & - & 27.5	&-	&-	&39.8 & \\
  &\textcolor{black}{RGRG}\cite{tanida2023interactive}&\textcolor{black}{\textbf{37.3}} &\textcolor{black}{\textbf{12.6}}&\textcolor{black}{\textbf{16.8}}&\textcolor{black}{26.4}&-&-&- \\
  &\textbf{Ours}&35.6 &11.1&14.1&26.5&\textbf{42.8}&\textbf{31.5}&\textbf{33.1} \\
	\midrule
 % IU-Xray&R2GenCMN\cite{chen2022crossmodal} &0.470 &0.165	&0.187	&0.371	&-	&-	&- &\\
 %  &Constrastive\cite{liu-etal-2021-contrastive} &0.492 &0.169	&0.193	&0.381	&-	&-	&- &\\
 %  &KiUT\cite{huang2023kiut}                  &\textbf{0.525}  &\textbf{0.199}	&\textbf{0.242}	&\textbf{0.409}	&-	&-	&- &\\
  % &KiUT\cite{huang2023kiut}                    &0.525 &0.199 &0.242 &0.409&-	&-	&- &\\
 %  &\textbf{Ours }                            &0.408  &0.133	&0.175	&0.317	&-	&-	&- &\\
  % \bottomrule
	\end{tabular}
	\label{tab:report}
\end{table}

% %%%% clinical significance test %%%%%
% ==============Analysis radiologits 1
% P: 0.6351137619580722
% shaoyifu: 0.15023724653948545
% minhang: 0.7700215400757205
% macau: 0.13488161551098374
% mimic: 0.395853076486303
% ==============Analysis radiologits 2
% P: 0.3399044981308701
% shaoyifu: 0.29936329338603507
% minhang: 0.358541003833548
% macau: 0.2923082087179252
% mimic: 0.3089502363630715
% ==============Analysis radiologits 3
% P: 0.0311418939042252
% shaoyifu: 0.058032893959273496
% minhang: 0.1711306563278037
% macau: 0.002403624236254015
% mimic: 0.400220448285021
% \begin{table}
% 	\caption{Clinical significance validation. We present the validation results for each radiologist at each center and the overall result.}
%   % As there are three radiologists involved, the P threshold is set to 0.0167 (0.05/3).
% 	\centering
%     \begin{tabular}{cccccc}
% 		\toprule
% 		\cmidrule(r){1-5}
% 		 &C1 &C2 &C3 &C4 &Overall\\
% 		\midrule
%           R1 &0.150 &0.770 &0.135 &0.396 &0.635\\
%           R2 &0.299 &0.359 &0.292	&0.309 &0.340\\
%           R3 &0.058 &0.171 &\textbf{0.026} &0.400 &\textbf{0.062}\\
% 	   \midrule
% 	\end{tabular}
% 	\label{tab:report}
% \end{table}

%%%%%%%%%% Population characteristics %%%%%%%%%%
\begin{table}
	\caption{Population characteristics of our training and validation dataset.}
  % As there are three radiologists involved, the P threshold is set to 0.0167 (0.05/3).
	\centering
    \begin{tabular}{ccp{10cm}}
		\toprule
		\cmidrule(r){1-3}
		 Dataset &Split &Population characteristics\\
		\midrule
          \textbf{MIMIC-CXR}\cite{johnson2019mimic} & train, validation & The dataset was collected at the Beth Israel Deaconess Medical Center in Boston, MA between 2011 - 2016. \\
          \textbf{Padchest} \cite{padchest} & train & This dataset was collected from San Juan Hospital (Spain) between 2009 and 2017. The dataset encompasses six distinct positional views. 50.3\% images correspond to women and 49.7\% to men. The median age of the population was 70, with a standard deviation of 20. \\
          \textbf{CXR-AL14} \cite{CXR-AL14} & train & - \\
          \textbf{VinDr-CXR} \cite{nguyen2022vindr} & train & The dataset were retrospectively collected from the Hospital 108 and the Hanoi Medical University Hospital between 2018 - 2020, two of the largest hospitals in Vietnam. The median age of the population was 43.77. \\
          \textbf{ChestXray14} \cite{wang2017chestx}  &validation &The dataset is extracted from the clinical PACS database at National Institutes of Health Clinical Center and consists of ~60\% of all frontal chest x-rays in the hospital. The median age of population was 49 with standard deviation of 17. 56.5\% images correspond to men and 43.5\% to women.\\
          \textbf{ChestX-Det} \cite{lian2021structure} & train & It is a subset of ChestXray14. The median age of population was 51 with standard deviation of 16. 57.2\% images correspond to men and 42.8\% to women.\\
          \textbf{CheXmask} \cite{cheXmask2023heart} & train & This dataset is collected from six public dataset. We only use MIMIC-CXR, Padchest, and VinDr-CXR for training. \\
          \textbf{SIIM} \cite{siim} & train & - \\
          \textbf{In-house dataset} & train & The dataset consists of 2531 images which were collected from three hospitals in China. All the images are captured from the front view. \\
          \textbf{MS-CXR} \cite{boecking2022making} &validation & It is a subset of MIMIC-CXR. \\
          \textbf{RSNA Pneumonia} \cite{shih2019augmenting}  & validation & The dataset was collected from the 112,000-image public National Institutes of Health (NIH) CXR8 dataset. 56.9\% images correspond to man and 43.1\% to women. \\
          \textbf{JSRT} \cite{jsrt2000lung}  & validation &  This database were collected from 13 medical centers in Japan and one institution in the United States. The average age of patients with nodules was 60 years old. 51\% correspond to man and 49\% correspond to women.\\
          \textbf{MS-PS} & validation & It is a subset of MS-CXR. \\ 
        \midrule
	\end{tabular}
	\label{tab:demography}
\end{table}

\setcounter{figure}{0}
\renewcommand{\figurename}{Supplementary Figure}

% \begin{figure}[t]
% 	\centering
% 	%\fbox{\rule[-.5cm]{4cm}{4cm} \rule[-.5cm]{4cm}{0cm}}
%     \includegraphics[width=\linewidth]{report_instruc.png}
% 	\caption{Explanation of the Report Instruction Subset in DR-VQA. We extract diseases and their relevant descriptions from the original MIMIC reports and combine them to create concise instructions. The Disease Phrase Distribution includes 10 diseases along with the number of attributes related to each disease. The Disease Attribute Distribution provides an introduction to location attributes and severity attributes.}
% 	\label{fig:Attribute}
% \end{figure}

\begin{figure}[t]
	\centering
	%\fbox{\rule[-.5cm]{4cm}{4cm} \rule[-.5cm]{4cm}{0cm}}
    \includegraphics[width=0.9\linewidth]{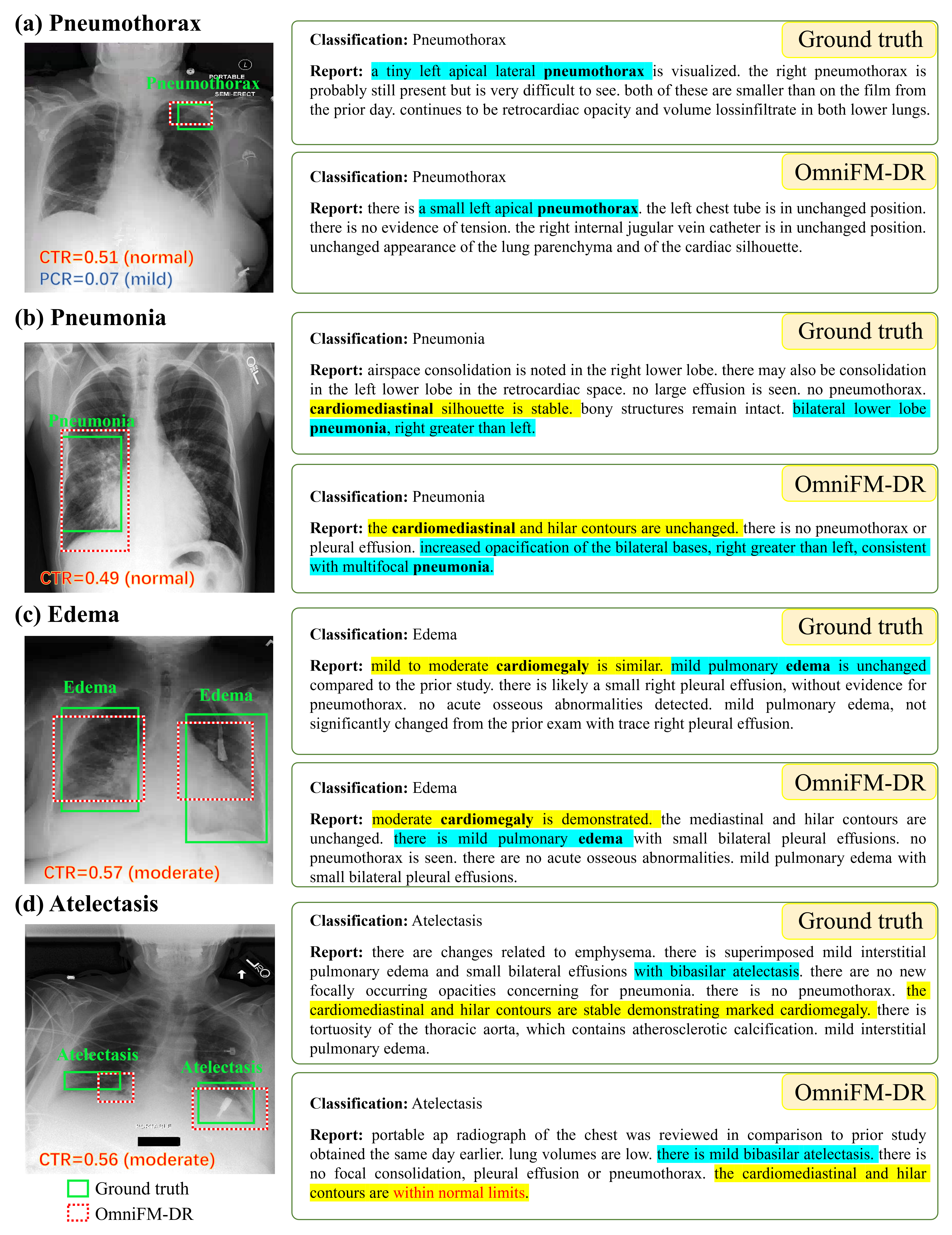}
 \caption{Typical success and failure cases of OmniFM-DR and ground truth for three tasks: multi-disease classification, disease localization, and report generation. (a) Pneumothorax; (b) Pneumonia; (c) Edema; (d) Atelectasis. In the left Chest X-ray image, the BBox with a green solid line denotes the ground truth, and the BBox with a red-white dashed line represents the region detected by OmniFM-DR. In the right reports, the blue highlighted text represents the matched classified lesions compared to the ground truth report, and the yellow highlighted area represents the matched report describing other categories (e.g. cardiomegaly). CTR and PCR denote the Cardiothoracic Ratio and Pneumothorax Compress Ratio, respectively.
 \textcolor{black}{For cases of (a) Pneumothorax, (b) Pneumonia, (c) Edema, our model is capable of predicting the classification, localization, and reporting of the lesions, and the results of each task are consistent. However, for the case of (d) Atelectasis, the AI exhibits deviation in localizing the atelectasis disease, and there is inconsistency in the description of increased cardiac silhouette in the report.}}
    \label{fig:examples}
\end{figure}

\begin{figure}[t]
	\centering
	%\fbox{\rule[-.5cm]{4cm}{4cm} \rule[-.5cm]{4cm}{0cm}}
    \includegraphics[width=0.95\linewidth]{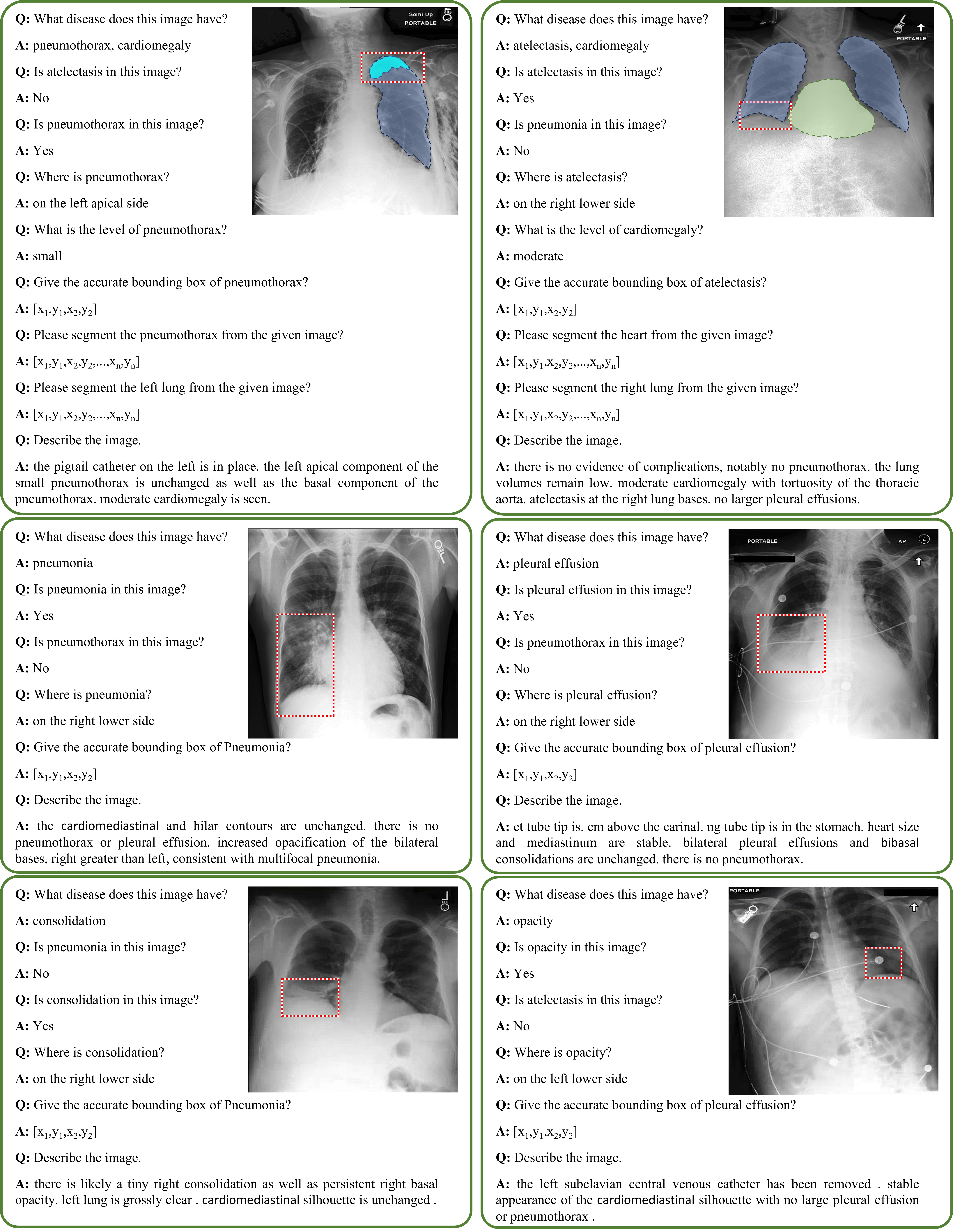}
	\caption{Typical examples of instruction set for six disease labels: Pneumothorax, Atelectasis, Pneumonia, Pleural Effusion, Condilidation, and Opacity. The left panel indicates the multiple instruction sets utilized during the training and testing phase. In the Chest X-ray image, the red dash line BBox denotes the region detected by OmniFM-DR. }
	\label{fig:instruction}
\end{figure}

% \begin{figure}[t]
% 	\centering
% 	%\fbox{\rule[-.5cm]{4cm}{4cm} \rule[-.5cm]{4cm}{0cm}}
%     \includegraphics[width=\linewidth]{labelPlatform.png}
% 	\caption{Overview of the annotation platform. Radiologists can review the pneumothorax contour, assess the accuracy of disease bounding boxes, and compare the differences between the original report and the AI-generated report all on the same page. Radiologists can click on the disease list on the left side to open a small window where they can provide feedback on missed detections, false positives, and descriptive errors related to each disease mentioned in the report.}
% 	\label{fig:labelPlatform}
% \end{figure}

\end{document}